\documentclass[11pt]{article}

\usepackage[final]{acl}

\usepackage{times}
\usepackage{wrapfig}
\usepackage{xcolor}         
\usepackage{color, colortbl}
\usepackage{latexsym}
\definecolor{MyPipelineYellow}{RGB}{254,228,160}
\definecolor{MyPipelineGreen}{RGB}{198,224,184}
\definecolor{MyPipelineBlue}{RGB}{191,215,237}
\usepackage[T1]{fontenc}
\usepackage{adjustbox}
\usepackage{multirow}\usepackage[utf8]{inputenc} 
\usepackage[T1]{fontenc}    
\usepackage{hyperref}       
\usepackage{url}            
\usepackage{booktabs}       
\usepackage{amsfonts}       
\usepackage{nicefrac}       
\usepackage{microtype}      
\usepackage{array}
\usepackage{wrapfig}
\usepackage{tabularx}
\usepackage{float}
\usepackage{xcolor}
\usepackage{pifont} 
\newcolumntype{C}[1]{>{\centering\arraybackslash}m{#1}}
\usepackage{xcolor}         
\usepackage{color, colortbl}
\definecolor{citecolor}{HTML}{2980b9}
\definecolor{linkcolor}{HTML}{c0392b}
\definecolor{darkorange}{HTML}{FF8C00}
\definecolor{chocolate}{HTML}{D2691E}
\definecolor{darkgreen}{HTML}{006400}
\definecolor{darkblue}{HTML}{00008B}
\definecolor{mediumblue}{HTML}{0000CD}
\definecolor{dodgerblue}{HTML}{1E90FF}
\definecolor{royalblue}{HTML}{4169E1}
\definecolor{shadecolor}{RGB}{237,237,237}
\definecolor{backred}{RGB}{255, 190, 190}
\definecolor{backblue}{RGB}{210, 230, 250}

\definecolor{zrrgreen}{HTML}{008000}
\definecolor{zrrblue}{HTML}{4682B4}
\definecolor{zrrred}{HTML}{B22222}

\definecolor{my_green}{RGB}{51,102,0}
\definecolor{my_red}{RGB}{204, 0, 0}
\newcommand{\cmark}{\textcolor{my_green}{\ding{51}}} 
\newcommand{\xmark}{\textcolor{my_red}{\ding{55}}} 
\newcolumntype{Y}{>{\centering\arraybackslash}p{3em}}

\newcommand{\ours}{SIV-Bench}
\definecolor{MyPipelineYellow}{RGB}{254,228,160}
\definecolor{MyPipelineGreen}{RGB}{198,224,184}
\definecolor{MyPipelineBlue}{RGB}{191,215,237}
\usepackage{amsmath}
\usepackage{amsthm}
\usepackage{amssymb}
\usepackage{graphicx}
\usepackage{subcaption}
\usepackage{multirow}
\usepackage{makecell}
\usepackage{caption}

\usepackage[utf8]{inputenc}

\usepackage{microtype}
\usepackage{subcaption}

\usepackage{inconsolata}

\usepackage{graphicx}

\title{SIV-Bench: A Video Benchmark for Social Interaction Understanding and Reasoning}

\author{%
  Fanqi Kong$^{1,2}$, Weiqin Zu$^{2,4}$, Xinyu Chen$^{1}$, Yaodong Yang$^{1}$, Song-Chun Zhu$^{1,2,3}$, Xue Feng$^{2}$\thanks{Corresponding Author.}\\ \\ 
  $^1$Peking University \quad \quad 
  $^2$State Key Laboratory of General Artificial Intelligence, BIGAI \\ 
    $^3$Tsinghua University \quad \quad $^4$ShanghaiTech University\\
    \texttt{kfq20@stu.pku.edu.cn, fengxue@bigai.ai}
}

\begin{document}
\maketitle
\begin{abstract}
Understanding social interaction, which encompasses perceiving numerous and subtle multimodal cues, inferring unobservable mental states and relations, and dynamically predicting others' behavior, is the foundation for achieving human-machine interaction. Despite rapid advances in Multimodal Large Language Models (MLLMs), the rich and multifaceted nature of social interaction has hindered the development of benchmarks that holistically evaluate and guide their social interaction abilities. Based on social relation theory, which has been widely regarded as a foundational framework for understanding social behavior, we provide SIV-Bench, a novel video benchmark for systematically evaluating MLLMs' capabilities across Social Scene Understanding (SSU), Social State Reasoning (SSR), and Social Dynamics Prediction (SDP).   SIV-Bench features 2,792 originally collected video clips and 5,455 meticulously generated question-answer pairs derived from a human-LLM collaborative pipeline. It covers 14 typical relationships, diverse video lengths, genres, presentation styles, and linguistic and cultural backgrounds. Our comprehensive experiments show that leading MLLMs perform relatively well on SSU but remain weak on SSR and SDP, with the systematic confusion in relation inference as a key bottleneck. An in-depth analysis of the reasoning process attributes MLLMs' suboptimal performance to misalignment with human thoughts and insufficient reasoning depth. Moreover, we find audio and subtitles aid in reasoning-intensive SSR and SDP. Together, SIV-Bench offers a unified testbed to measure progress, expose limitations, and guide future research toward more socially intelligent MLLMs. We release the dataset and code at our project website: \url{https://kfq20.github.io/sivbench/}.
\end{abstract}

\begin{figure*}[t!]
    \centering
    \includegraphics[width=0.99\linewidth]{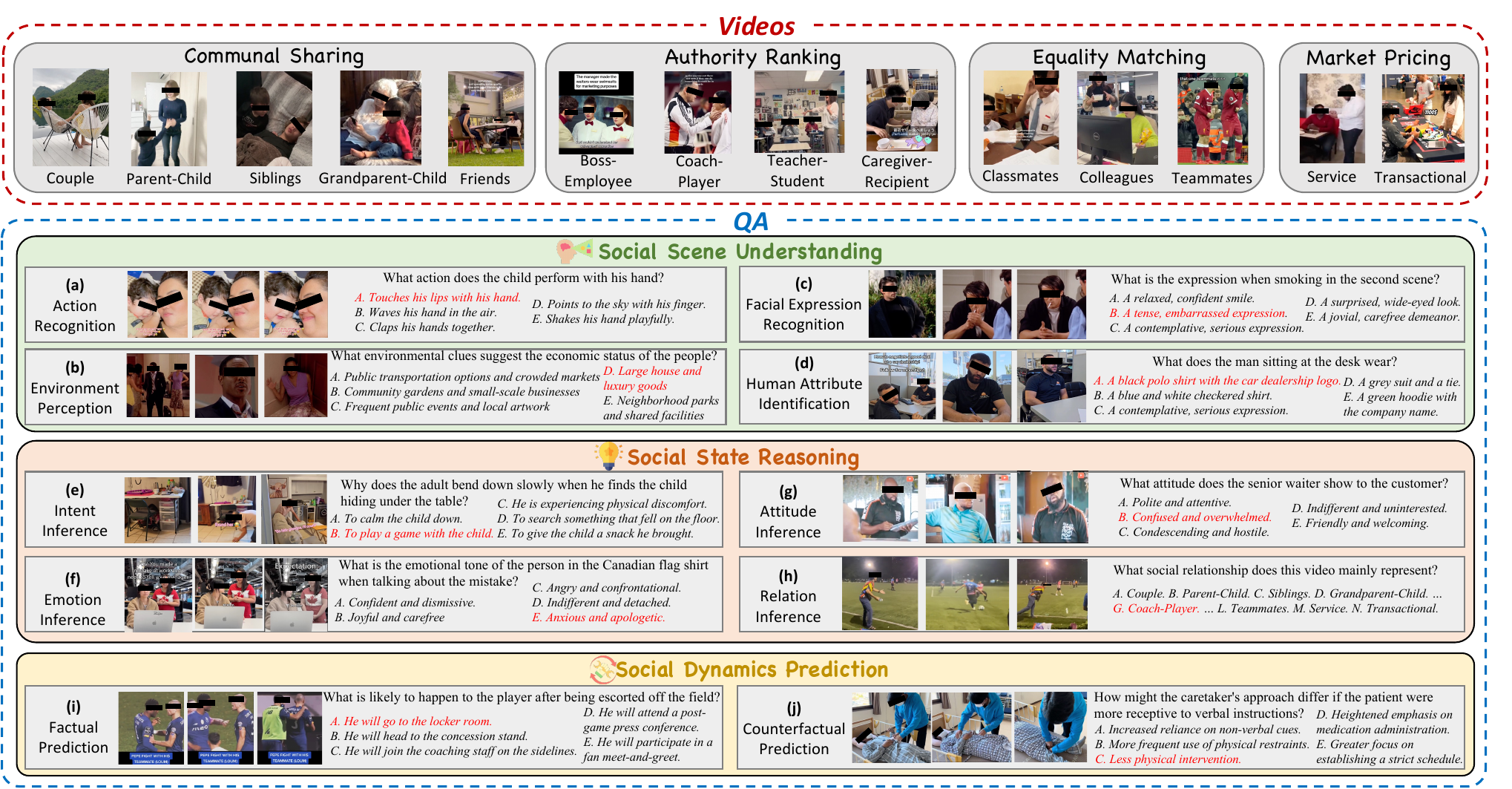}
    \caption{Overview of \ours{}, showing its diverse videos spanning various social interactions and sample QAs for three task dimensions: Social Scene Understanding (SSU), Social State Reasoning (SSR), and Social Dynamics Prediction (SDP), along with their fine-grained sub-tasks.}
    \label{fig:main}
    \vspace{-0.3cm}
\end{figure*}

\section{Introduction}
The rapid development of Multimodal Large Language Models (MLLMs) capable of processing text, images, and video has driven strong performance across tasks such as visual reasoning, video captioning, and multimodal dialogue \citep{team2023gemini, wu2024next, zhu2025internvl3exploringadvancedtraining, hurst2024gpt, yang2024qwen2, wu2026autowebworldsynthesizinginfiniteverifiable}. As these capabilities expand, there is a growing need for benchmarks that can evaluate model performance, uncover limitations, and guide future research \citep{fu2024video, fang2024mmbench, zhou2024mlvu, qiang2025mledojointeractiveenvironmentsempowering, huang2024adasociety, Li_2023_ICCV, xie2026visjudgebenchaestheticsqualityassessment}. One critical yet underexplored area is the social interaction understanding and reasoning, which is a core aspect of social intelligence that encompasses not only observable behaviors but also implicit mental states and social relationships governing behaviors such as forming bonds, exchanging information, and coordinating actions \citep{berger1972status, smith1988analyzing}. 
However, existing video benchmarks, whether designed for specific tasks like video object segmentation \citep{ding2025mosev2, athar2025vicas}, captioning \citep{chen2025vidcapbench, wu2025event}, and fine-grained action understanding \citep{perrett2025hd}, or for broader video understanding \citep{wang2025lvbench, li2024mvbench, li2024videovista}, still struggle to systematically probe MLLMs' understanding of the multifaceted nature of social interaction. 

To address this gap, we \textit{first} decompose the capacity to understand and reason about social interaction into three core, interrelated dimensions. \textbf{1) Social Scene Understanding (SSU)} is foundational, enabling the recognition of visible elements such as objects, environments, and socially salient human features like body movements, clothing, and physical appearance.  Reliable scene perception is required to ground interpretations in relevant cues. \textbf{2) Social State Reasoning (SSR)} is essential for interpreting the unobservable states of interaction, such as emotions, intents, attitudes, and interpersonal relationships, which guide and shape behavior \citep{strachan2024testing, wu2020mentalizing, kong2025enhancing}. This capacity allows models to move beyond surface-level features and grasp the underlying states. \textbf{3) Social Dynamics Prediction (SDP)} enables the model to reason about how interactions proceed over time or under alternative conditions, capabilities essential for a flexible and human-like understanding of social scenarios \citep{ramnani2004system, byrne2016counterfactual}. It involves factual prediction and counterfactual prediction. The former anticipates upcoming actions or changes in social states, while the latter examines how alterations in social scenes or states can affect interaction outcome.

\textit{Second}, to systematically and comprehensively evaluate the three dimensions, we introduce \textbf{\ours{}} (\textbf{S}ocial \textbf{I}nteraction \textbf{V}ideo \textbf{Bench}mark) grounded in social relation theory, which has been widely regarded as a foundational framework for understanding social behavior \citep{thibaut2017social, burkitt1997social, hartup1989social}. Specifically, SIV-Bench is built on Fiske's Relational Models Theory \citep{fiske1992four}, categorizing social interactions via four foundational models (\textit{Communal Sharing}, \textit{Authority Ranking}, \textit{Equality Matching}, and \textit{Market Pricing}), instantiated through 14 relation types (e.g., parent-child, friends, colleagues). This relational context underpins all three dimensions. It conditions how cues are interpreted in SSU (e.g., a gaze between colleagues vs. lovers), modulates mental-state inference in SSR (e.g., criticism from a mentor vs. a stranger), and shapes SDP by constraining future behaviors and counterfactuals through relational norms and history (e.g., siblings vs. business rivals). SIV-Bench comprises 2,792 video clips sourced from TikTok and YouTube, and a QA pipeline that combines adversarial filtering with human verification produces 5,455 high-quality questions. To systematically assess the contribution of linguistic cues, SIV-Bench further provides audio tracks and three subtitle conditions: the original version (origin), a version with transcribed dialog added (+sub), and a version with all on-screen text removed (-sub).

\begin{figure*}[t!]
    \centering
    \includegraphics[width=0.99\linewidth]{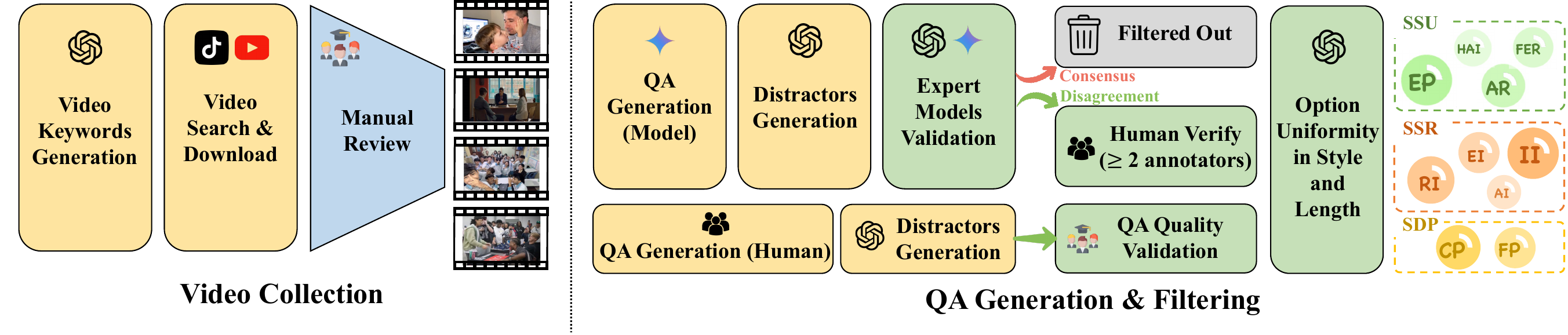}
    \caption{The \ours{} construction pipeline, detailing the data collection process (left), and the QA generation \& filtering process (right) with human-LLM collaboration. In the diagram, \textcolor{MyPipelineYellow}{\rule{0.8em}{0.8em}} blocks indicate content (like keywords, video and QA) generation steps, \textcolor{MyPipelineGreen}{\rule{0.8em}{0.8em}} blocks represent validation or modification stages.}
    \label{fig:qa pipeline}
\end{figure*}

In the experiments, we evaluate a broad set of models, including leading commercial MLLMs, strong open-source MLLMs, and video-specialized models. Overall, models are comparatively strong on SSU but weak on high-level reasoning, particularly within SSR, where \textbf{relation inference} exhibits systematic confusions. Our fine-grained analyses reveal recurring error sources, including selecting plausible but secondary relations, over-reliance on contextual cues, insufficient commonsense social reasoning, and missed perceptual details. SDP remains challenging as well, with counterfactual cases being relatively better handled by top models. We further study textual cues via controlled subtitle ablations and observe a task-dependent effect. Removing language typically has a limited impact on basic perception, but can degrade performance on inference-intensive SSR/SDP cases. Furthermore, on a hard diagnostic subset with brief explanations and a human baseline, we observe a large remaining gap to humans, underscoring the difficulty of robust, human-aligned social reasoning.

Our key contributions are as follows: \textbf{1)} We propose a novel analytical framework that structurally decomposes the complex task of multimodal social interaction understanding and reasoning into three core, interrelated dimensions, each further detailed into fine-grained sub-tasks. \textbf{2)} We introduce \ours{}, a new video benchmark specifically curated for the analysis and comprehension of complex real-world social interactions. \ours{} comprises 2,792 real-world video clips representing 14 distinct social relationship types, and features 5,455 high-quality question-answer pairs generated through a human-LLM collaborative pipeline. \textbf{3)} Our comprehensive experiments on diverse MLLMs reveal the limitations in their current capacity for deep human social understanding.

\section{\ours}
This section presents the construction of \ours{}, including video collection (Section~\ref{sec: video collect}), QA composing (Section~\ref{sec: qa generation}), and a comparison with existing benchmarks (Section~\ref{sec: comparison}). Figure \ref{fig:main} offers some illustrative examples from \ours{}. Figure \ref{fig:qa pipeline} shows the construction pipeline.

\subsection{Video Collection} \label{sec: video collect}
Firstly, we utilize GPT-4o-mini \citep{hurst2024gpt} to generate comprehensive search keywords for each of the 14 relationship types (the word-clouds are shown in Figure \ref{fig:keywords_cloud}), specifically including terms associated with varying degrees of intimacy, both positive (e.g., "love", "encourage") and negative (e.g., "conflict", "fight"). Leveraging these keywords, we conduct targeted searches and download initial video candidates from TikTok and YouTube platforms using Python libraries such as \texttt{TikTokApi} and \texttt{yt-dlp}, yielding approximately 5000 raw video clips. Each video is then manually reviewed by the authors to ensure it contains clearly observable and meaningful social interactions. Videos are excluded if they do not depict clear social interaction (e.g., a vlogger speaking directly to the camera without interacting with others), if the interaction context does not fit within a set of well-defined interpersonal scenarios (e.g., an interview setting with scripted dialogue), or if the dominant interaction is difficult to identify due to the presence of multiple overlapping social dynamics (e.g., a large multi-generational family posing for a group photo). These criteria are designed to ensure that each included video primarily features one interpretable and coherent type of interpersonal interaction, allowing for more consistent analysis.

\begin{figure}[ht]
    \centering
    \begin{subfigure}[t]{0.48\columnwidth}
        \centering
        \includegraphics[width=\linewidth]{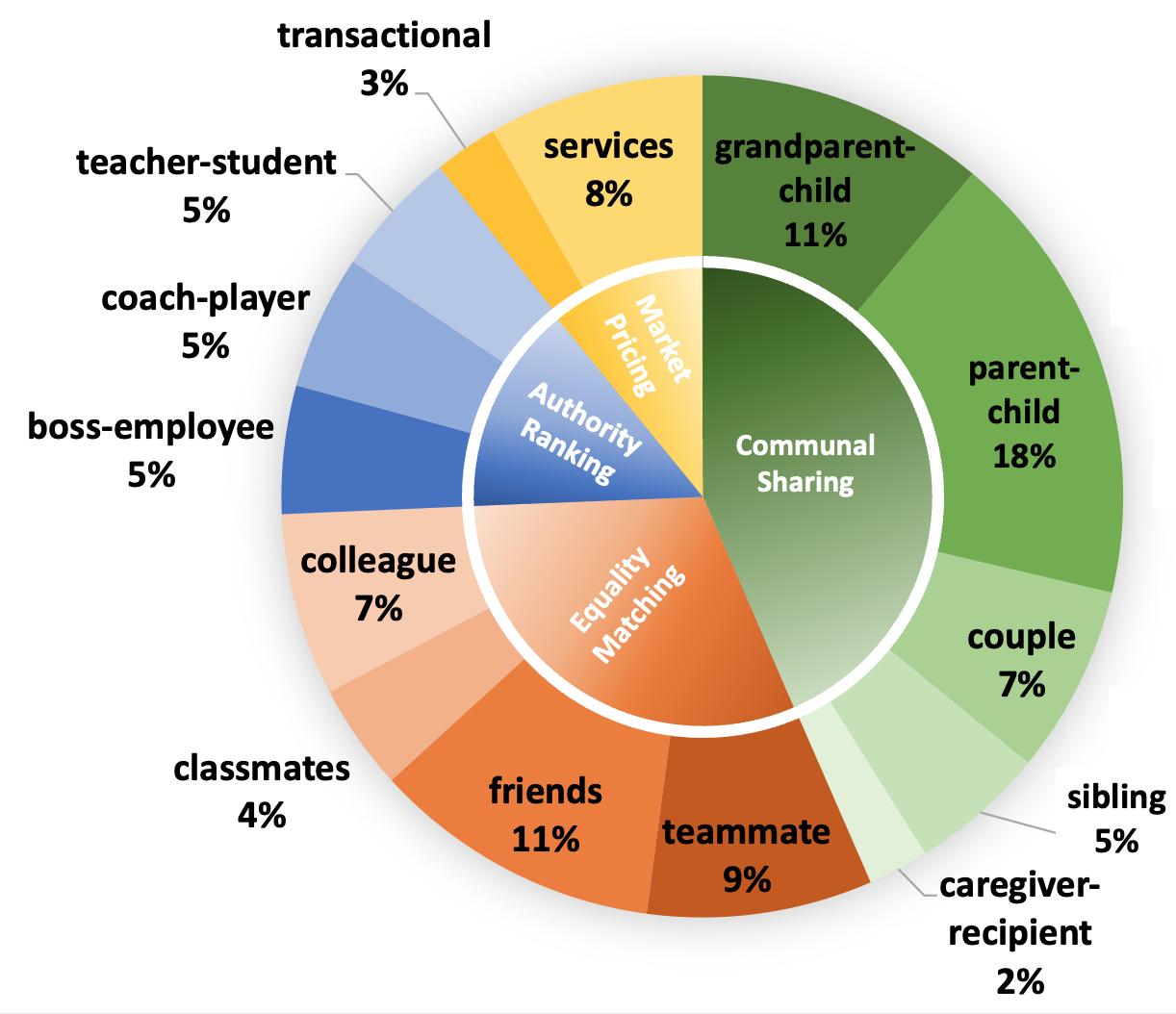}
        \caption{Relation type distribution.}
        \label{fig:relation num}
    \end{subfigure}\hfill
    \begin{subfigure}[t]{0.48\columnwidth}
        \centering
        \includegraphics[width=\linewidth]{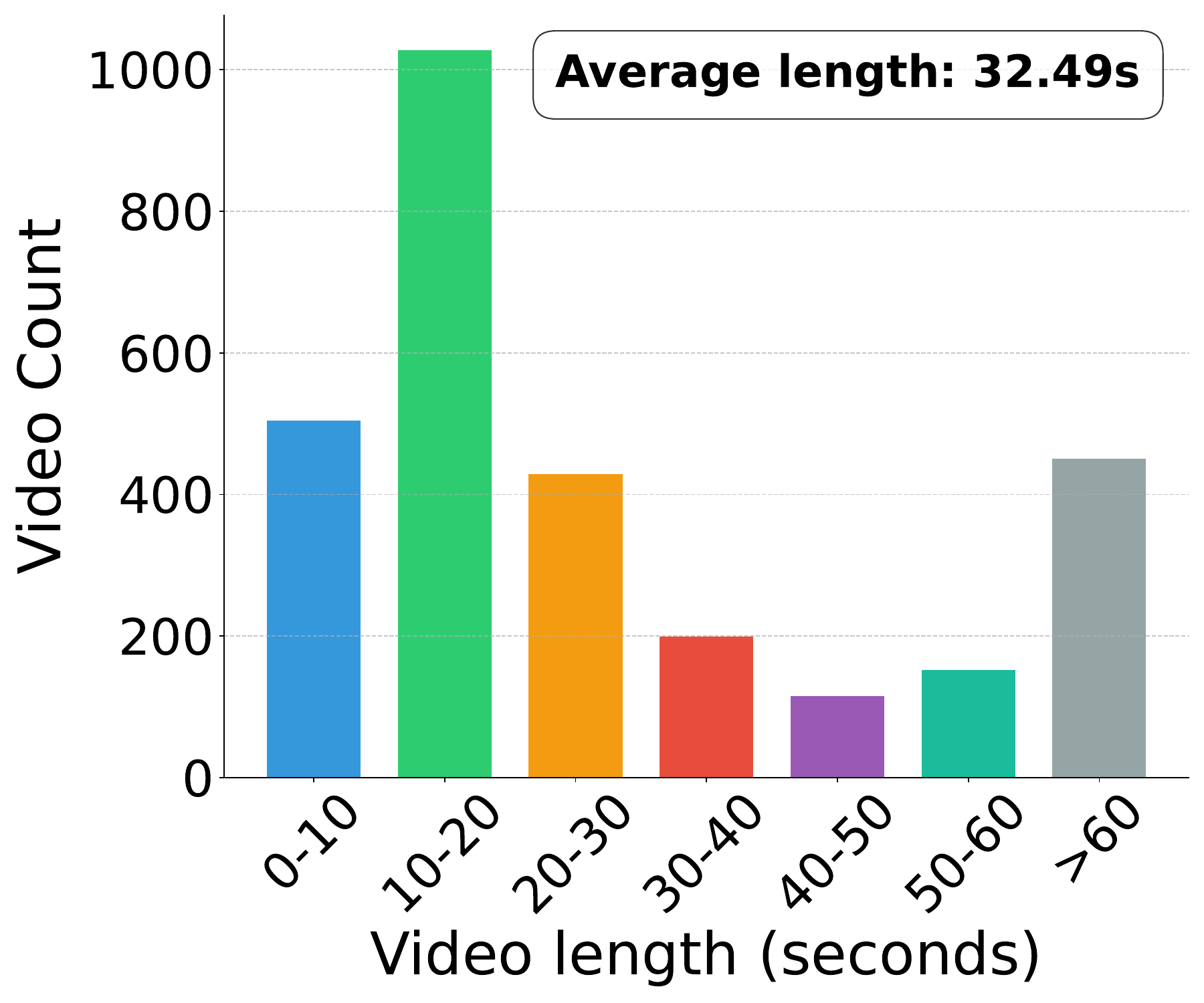}
        \caption{Video length distribution.}
        \label{fig:video length}
    \end{subfigure}
    \caption{Video statistics for \ours{}.}
    \label{fig:video stats}
    \vspace{-0.2cm}
\end{figure}

The \ours{} comprises \textbf{2,792} curated video clips, with statistics shown in Figure \ref{fig:video stats}. The collection showcases a rich diversity in social relationships (Figure \ref{fig:relation num}). Communal Sharing interactions are the most represented category, reflecting the naturalistic prevalence and psychological centrality of these relationships in daily life \citep{simao2014gratitude, kameda2005social}. The other three relational models are also well represented, ensuring broad interpersonal coverage. In addition to relationship diversity, the benchmark covers heterogeneous genres and filming/editing conventions, which helps test models under different visual narratives and interaction realizations. In terms of duration (Figure~\ref{fig:video length}), clips average 32.49 seconds. While most clips are 10–20 seconds, the dataset spans a wide distribution, including many short clips (under 10 seconds) and a significant number over 60 seconds. English predominates, but \ours{} also includes other languages, adding to multicultural diversity (detailed in Figure \ref{fig: language}). Details on video diversity are in Appendix \ref{appdx: video diversity}.

\begin{figure}[htbp]
\centering
\includegraphics[width=0.4\textwidth]{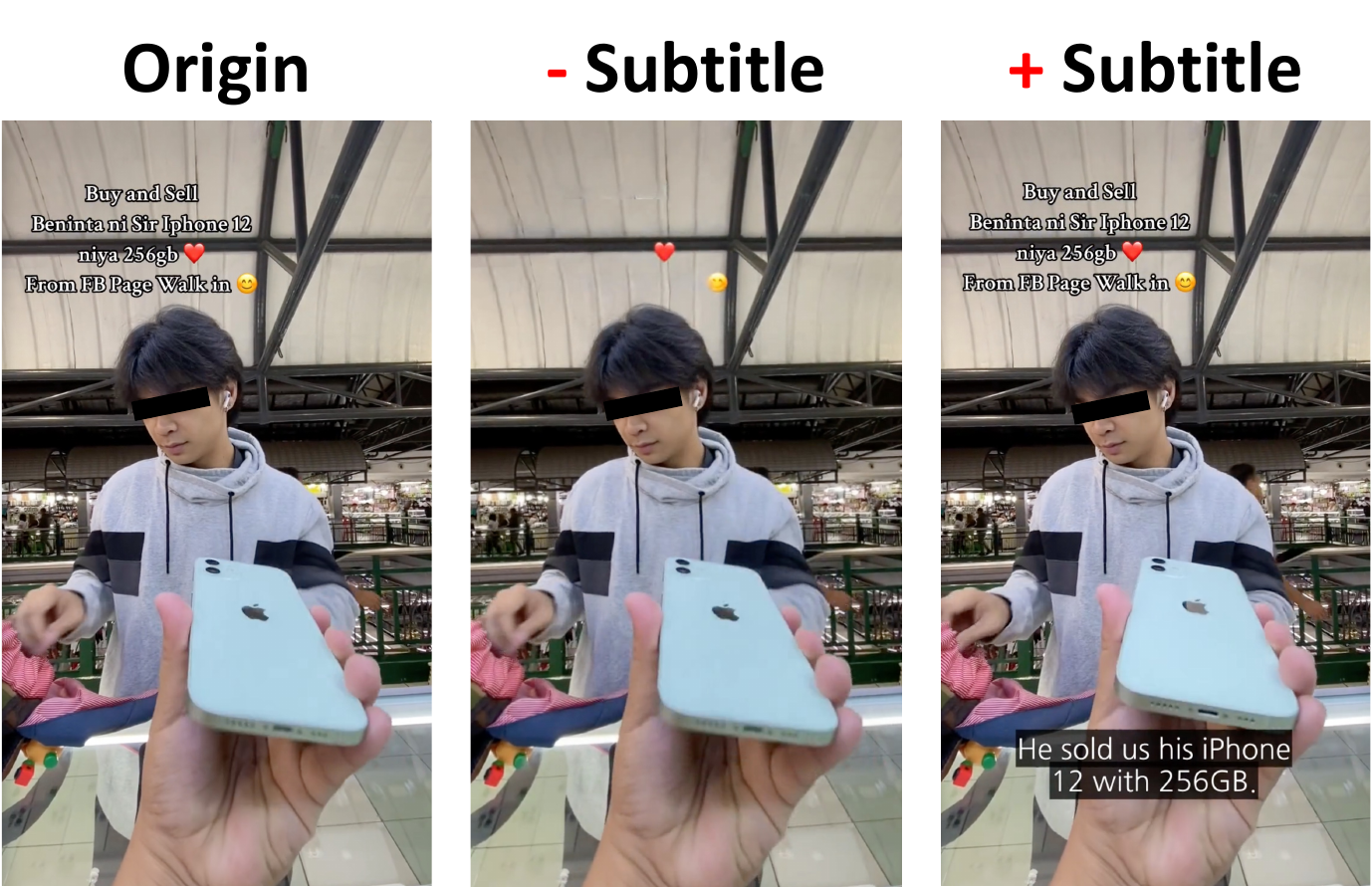}
\caption{Illustration of the three subtitle conditions.}
\label{fig:subtitle}
\vspace{-0.3cm}
\end{figure}

To evaluate how different forms of textual information affect MLLMs' understanding of social interactions, we implement specific subtitle processing methods (Figure~\ref{fig:subtitle}). Many original videos (`Origin') contain embedded on-screen text that often serves as scene descriptors or keywords (e.g., the `Buy and Sell' text overlay). To focus on visual and auditory cues, we create a `-Subtitle' version by removing such original textual overlays using \texttt{video-subtitle-remover} \citep{yaofanguk_video_subtitle_remover}. Conversely, to provide full access to spoken dialogue, we generate a `+Subtitle' version. We employ Whisper-large-v3 \citep{radford2022whisper} for audio transcription of the dialogue, and then use GPT to translate these transcriptions into English, ensuring consistent and high-quality subtitles (e.g., `He sold us his iPhone 12 with 256GB'). 

\begin{table*}[t!]
\centering

\caption{Comparison of various benchmarks, including total number of items (\textbf{\#Items}, representing the number of videos, dialogues, images, etc.), number of QA pairs (\textbf{\#QAs}), annotation method (\textbf{Anno.}, M/A means manually/automatic manner), and tasks (\textit{Task Types}: \textbf{SU} for Scene Understanding, \textbf{SR} for State Reasoning, \textbf{DP} for Dynamics Prediction). Note that \textit{Task Types} here refer to general task categories, which include both social and physical scenarios. The table also shows whether each benchmark includes \textbf{Multi-Person Interaction}, covers \textbf{Various Relations}, is based on newly collected data (\textbf{Original Collection}), and provides subtitle/audio (\textbf{S.A.}).}
\vspace{5pt}
\begin{adjustbox}{max width=0.98\textwidth}
\begin{tabular}{lcccccccccc}
\toprule
\multirow{2}{*}{\textbf{Benchmark}} & \multirow{2}{*}{\textbf{\#Items}} & \multirow{2}{*}{\textbf{\#QAs}} & \multirow{2}{*}{\textbf{Anno.}}&  \multicolumn{3}{c}{\textit{\textbf{Task Types}}} & \textbf{Multi-Person} & \textbf{Various} & \textbf{Original} & \multirow{2}{*}{\textbf{S.A.}} \\
\cmidrule(lr){5-7}
& & & & \textbf{SU} & \textbf{SR} & \textbf{DP} & \textbf{Interaction} & \textbf{Relations} & \textbf{Collection} &  \\
\midrule
\multicolumn{11}{c}{\textit{Social Relation Inference}} \\
\midrule
DialogRE \citep{yu2020dialogue} & 1,788 & - & M & \cmark & \cmark & \xmark & \cmark & \cmark & \xmark &  - \\
PIPA \citep{sun2017domain} & 37,107 & - & M & \xmark & \cmark & \xmark & \cmark & \cmark & \xmark &  - \\
ViSR \citep{Liu2019Social} & 8,000 & - & M & \xmark & \cmark & \xmark &  \cmark & \cmark & \cmark &  \xmark \\
\midrule
\multicolumn{11}{c}{\textit{General Video Understanding and Reasoning}} \\
\midrule
VideoMME \citep{fu2024video} & 900 & 2,700 & M & \cmark & \cmark & \xmark & \cmark & \xmark & \cmark &  \cmark \\ 
MLVU \citep{zhou2024mlvu}& 1,730 & 3,102 & M & \cmark & \xmark & \cmark & \cmark & \xmark & \xmark &  \xmark \\
VideoVista \citep{li2024videovista} & 894 & 24,906 & A & \cmark & \xmark & \cmark & \xmark & \xmark & \xmark &  \cmark\\
Social-IQ 2.0 \citep{siq2}& 1,000 & 6,000 & M & \cmark & \cmark & \xmark & \cmark & \cmark & \cmark &  \xmark \\
Social Genome \citep{mathur2025social} & 272 & 1,486 & M & \cmark & \cmark & \xmark & \cmark & \xmark & \xmark & \cmark \\
Perception Test \citep{patraucean2023perception}& 11,620 & 38,000 & M & \cmark & \xmark& \cmark & \xmark & \xmark & \cmark &  \xmark\\
MVBench \citep{li2024mvbench}& 3,641 & 4,000 & A & \cmark & \xmark & \cmark &  \xmark & \xmark & \xmark & \xmark \\
Video-Bench \citep{ning2023video}& 5,917 & 17,036 & A\&M & \cmark & \cmark &\xmark & \cmark & \xmark & \xmark &  \xmark\\
EgoSchema \citep{mangalam2023egoschema}& 5,063 & 5,063 & A\&M & \cmark & \cmark & \xmark & \cmark & \xmark & \xmark &  \xmark \\
\midrule
\ours{} & 2,792 & 5,455 & A\&M & \cmark & \cmark & \cmark & \cmark & \cmark & \cmark  & \cmark\\
\bottomrule
\end{tabular}
\end{adjustbox}

\label{tab:comparison}
\vspace{-0.2cm}
\end{table*}

\subsection{QA Composing} \label{sec: qa generation}
Our Question-Answer (QA) composition pipeline is designed to move beyond simple recognition tasks and ensure a high density of challenging reasoning problems. The process consists of three main stages: scalable generation, adversarial filtering, and human-in-the-loop curation. 

\textbf{Scalable Generation.} We begin by generating a large initial pool of diverse QA pairs for each video, leveraging the capabilities of Gemini-2.0-Flash with full video input. The prompts are carefully designed to elicit questions that span our SSU, SSR, and SDP evaluation dimensions. This stage typically yields an average of 10 QA pairs per video. Subsequently, we employ GPT-4o-mini to generate four distractors for each QA pair, ensuring they are contextually relevant yet clearly distinguishable from the correct answer. Separating distractor generation significantly reduces ambiguity compared to simultaneous generation.

\textbf{Adversarial Filtering.} To ensure \ours{} serves as a rigorous diagnostic tool rather than a saturated test set, we implement a strict \textit{Model-Consensus Filtering} mechanism. We utilize three commercial models (Gemini-2.0-Flash, Gemini-2.0-Pro, and GPT-4o-mini) to independently answer all candidate questions.
We adopt an adversarial approach: questions correctly answered by all models (indicating trivial reasoning or visual obviousness) are discarded. This can also prevent the model from choosing the correct answer without looking at the question itself or the video prompt~\citep{li2025vegas}. We focus exclusively on the remaining "hard" samples where model consensus is not achieved, as these disagreements signal the need for deeper social reasoning.

\begin{figure}[htbp]
\centering
\includegraphics[width=0.48\textwidth]{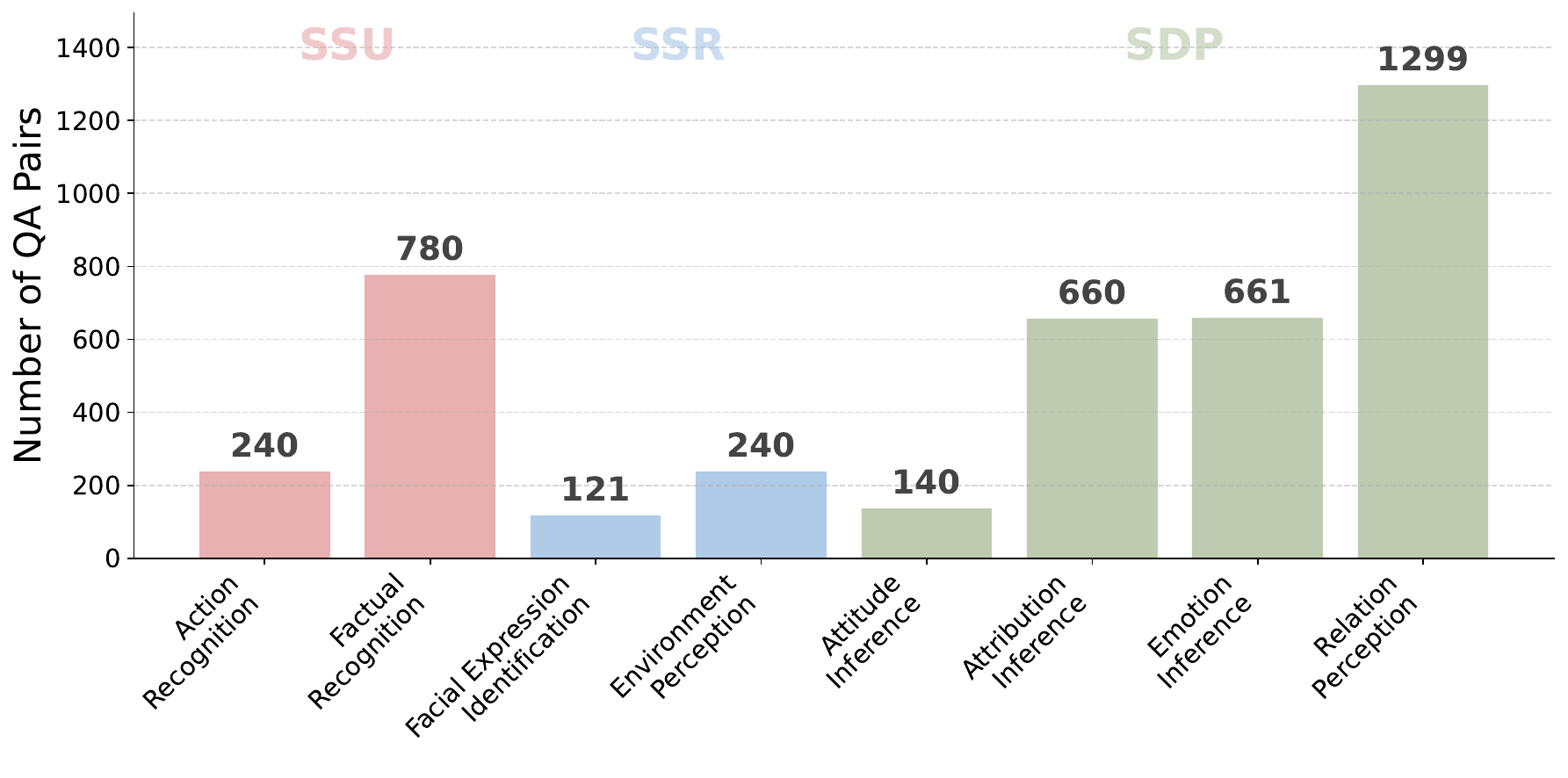}
\caption{QA Distribution across 10 sub-tasks.}
\label{fig:qa stats}
\vspace{-0.2cm}
\end{figure}

\textbf{Human Verification and Authoring.} The questions surviving the adversarial filter undergo a rigorous human-in-the-loop process to distinguish legitimate challenging samples from noise: \textbf{(1)} We recruit 20 human annotators to verify the non-consensus items. Each QA is reviewed by at least two annotators. Only questions where humans independently agree on the correct answer are retained. This yields 3,096 high-quality, model-challenging QAs. \textbf{(2)} To further expand the benchmark's ceiling, annotators are instructed to generate novel, complex questions and their answers. These human-authored 2,359 QAs are designed to probe nuances often missed by model generators. Appendix \ref{appdx: human} details the annotation guidelines and interface. For all new human-generated QAs, distractors are created using GPT-4o-mini. 

Finally, all curated QA pairs undergo an automated refinement stage to standardize linguistic style and option length, minimizing superficial cues that models might exploit. Full prompts are provided in Appendix \ref{appdx: qa generation}. Through this "Filter-then-Verify" pipeline, \ours{} ultimately comprises \textbf{5,455} high-quality QA pairs. The distribution across core dimensions is illustrated in Figure \ref{fig:qa stats}, featuring a strong emphasis on SSR and SDP, enabling rich evaluation of higher-order social intelligence. Appendix \ref{appdx: qa stats} provides detailed statistics confirming structural balance and diversity.

\subsection{Comparison with Existing Benchmarks} \label{sec: comparison}
Table \ref{tab:comparison} highlights the key differences between \ours{} and existing works. 
Prior datasets on social interaction, such as DialogRE \citep{yu2020dialogue} (text-only), PIPA \citep{sun2017domain} (image-only), and ViSR \citep{Liu2019Social}, are limited to relation recognition, single modalities, and lack task diversity and scalability due to manual annotation.
Currently, most general video understanding benchmarks (like VideoVista \citep{li2024videovista} and MVBench \citep{li2024mvbench}) lack a focus on social interaction; others (like MLVU \citep{zhou2024mlvu} and Video-Bench \citep{ning2023video}), while touching upon it, do not fully cover social relations. Even Social-IQ 2.0 \citep{siq2}, which concentrates on this area, has limitations in task diversity and dynamic reasoning. \ours{} is built on original data, combines manual and automatic annotations, and is one of the few benchmarks to provide subtitle and audio information, supporting richer multimodal social reasoning.

\begin{table*}[t!]
\centering
\caption{Evaluation results of MLLMs on SIV-Bench. Accuracy (\%) on SSU, SSR, SDP, and Overall under different subtitle settings (`origin', `+sub', `-sub'). Statistical significance tests are reported in Appendix \ref{app:statistical_analysis}.}
\vspace{5pt}
\begin{adjustbox}{max width=\textwidth}
\begin{tabular}{lc*{12}{Y}}
\toprule
\multicolumn{1}{c}{\multirow{2}{*}{\textbf{Models}}} & \multirow{2}{*}{\textbf{Params}} & \multicolumn{3}{c}{\textbf{Social Scene Understanding}} & \multicolumn{3}{c}{\textbf{Social State Reasoning}} & \multicolumn{3}{c}{\textbf{Social Dynamics Prediction}} & \multicolumn{3}{c}{\textbf{Overall}}\\
\cmidrule(lr){3-5} \cmidrule(lr){6-8}  \cmidrule(lr){9-11}  \cmidrule(lr){12-14}  
& & origin& + sub & - sub & origin & + sub & - sub & origin& + sub & - sub & origin & + sub & - sub\\
\midrule
\multicolumn{14}{c}{\textit{Open-source MLLMs}} \\
\midrule
mPLUG-Owl3               & 7B  & 46.11 & 45.94 & 46.34 & 39.78 & 39.50 & 38.13 & 44.30 & 46.08 & 44.35 & 42.06 & 42.42 & 41.15 \\
LLaVA-OneVision           & 7B  & 39.20 & 39.41 & 40.08 & 41.95 & 43.65 & 38.66 & 43.79 & 44.51 & 39.42 & 41.97 & 43.04 & 39.42 \\
LLaVA-Video               & 7B  & 50.22 & 50.61 & 50.56 & 39.33 & 38.19 & 36.14 & 41.60 & 42.14 & 39.94 & 41.09 & 42.00 & 38.66 \\
Qwen2.5-VL-7B-Instruct    & 7B  & 51.22 & 50.88 & 50.22 & 40.24 & 38.94 & 37.66 & 42.69 & 43.82 & 42.24 & 44.02 & 44.21 & 41.65 \\
InternVL3-8B              & 8B  & 56.83 & 56.13 & 56.50 & 40.35 & 40.90 & 37.92 & 44.53 & 45.52 & 44.40 & 45.82 & 46.05 & 44.56 \\
Qwen2.5-VL-72B-Instruct   & 72B & 75.73 & 76.24 & 73.54 & \underline{52.25} & \underline{52.75} & \underline{51.21} & 59.02 & 58.40 & 57.78 & \underline{58.80} & \underline{59.63} & \underline{57.66} \\
InternVL3-78B             & 78B & 71.46 & 73.66 & 71.76 & 51.65 & 52.39 & 50.14 & 55.77 & 56.28 & 54.25 & 55.46 & 56.32 & 54.50 \\
\midrule
\multicolumn{14}{c}{\textit{Closed-source MLLMs}} \\
\midrule
o4-mini                   & -   & 78.83 & 79.04 & 78.13 & 50.47 & 51.30 & 48.99 & 56.89 & 56.00 & 55.26 & 55.68 & 55.89 & 54.54 \\
GPT-4o                    & -   & 79.10 & 79.74 & 78.06 & 52.73 & 53.20 & 51.79 & 59.02 & \underline{60.59} & \underline{58.60} & 58.02 & 58.86 & 56.99 \\
Gemini-2.0-Flash          & -   & 78.46 & 78.16 & 78.34 & 51.89 & 52.43 & 49.78 & 57.59 & 58.63 & 55.70 & 56.40 & 57.23 & 54.64 \\
Gemini-2.5-Flash          & -   & \underline{81.70} & \underline{82.14} & \underline{79.71} & 48.99 & 50.54 & 47.60 & \underline{59.47} & 59.95 & 56.88 & 57.87 & 58.11 & 56.05 \\
Gemini-2.5-Pro            & -   & \textbf{85.07} & \textbf{85.41} & \textbf{84.94} & \textbf{54.30} & \textbf{54.85} & \textbf{52.32} & \textbf{60.45} & \textbf{61.54} & \textbf{58.83} & \textbf{61.65} & \textbf{62.40} & \textbf{60.22} \\
\bottomrule

\end{tabular}
\end{adjustbox}
\label{tab:main results}
\end{table*}

\section{Experiments}
\subsection{Experimental Setup}
We evaluate a diverse set of closed- and open-source MLLMs on \ours, including Gemini-2.0/2.5-Flash \citep{GoogleGemini2FlashDevBlog2025, Google2025Gemini2.5FlashBlog}, Gemini-2.5-Pro \citep{GoogleGemini2.5ProUpdateMay2025}, GPT-4o \citep{hurst2024gpt}, o4-mini \citep{OpenAISystemCardO4mini2025}, Qwen2.5-VL-7B/72B-Instruct \citep{bai2025qwen2}, mPLUG-Owl3 \citep{ye2024mplug}, InternVL3-8B/78B \citep{zhu2025internvl3}, LLaVA-OneVision \citep{li2024llava}, and LLaVA-Video \citep{zhang2024videoinstructiontuningsynthetic}. Evaluations are conducted using VLMEvalKit \citep{duan2024vlmevalkit}, which provides a unified evaluation interface across MLLMs. All models generate responses under their default inference settings to ensure a fair comparison.

We use a standardized prompt (Figure~\ref{fig:evaluate prompt}) across all models, asking for both an option letter (e.g., `A.') and the full answer text. To parse outputs robustly, we first extract a valid option letter; if missing, we match the raw output to all options using text similarity. To enable scalable and model-agnostic evaluation across heterogeneous MLLMs, we cast all tasks into a unified multiple-choice interface and report accuracy as the primary metric.

\subsection{Overall Performance}

The results shown in Table~\ref{tab:main results} highlights a clear performance stratification primarily driven by model scale. While the proprietary Gemini-2.5-Pro achieves the state-of-the-art (62.40\% w/ subtitles), large-scale open-source models demonstrate exceptional competitiveness; notably, Qwen2.5-VL-72B (59.63\%) outperforms other leading proprietary systems like GPT-4o and Gemini-2.5-Flash (>58\%). A significant capacity gap is evident, as larger models consistently dominate their smaller counterparts (e.g., Qwen-72B's 59.63\% vs. 7B's 44.21\%), suggesting that sufficient parameter count is a prerequisite for robust social reasoning. We verify that the observed performance gaps are statistically significant using paired bootstrap tests; full details are reported in Appendix \ref{app:statistical_analysis}.

\subsection{Decomposed Performance}
\begin{table}[ht]
\centering
\caption{Relative performance change (\%) against the baseline. The audio ablation (w/o Audio) is evaluated on Gemini-2.5-Flash, while subtitle variations (+Sub/-Sub) reflect the average impact across all models.}
\label{tab:ablation}
\resizebox{0.8\linewidth}{!}{%
\begin{tabular}{l|ccc|c}
\toprule
\textbf{Condition} & \textbf{SSU} & \textbf{SSR} & \textbf{SDP} & \textbf{Overall} \\
\midrule
w/o Audio & -0.35 & -2.44 & -2.33 & -1.40 \\
\midrule
+ Subtitle & +0.05 & +0.28 & +0.65 & +0.15 \\
- Subtitle & -0.97 & -2.07 & -1.68 & -0.90 \\
\bottomrule
\end{tabular}%
}
\end{table}

\textbf{Subtitle and Audio Influence.} To assess the impact of different modalities, we analyze the relative performance changes compared to the original video baseline, as shown in Table \ref{tab:ablation}. The audio ablation is conducted on Gemini-2.5-Flash, while subtitle variations represent the average results across all evaluated models. The results demonstrate a clear hierarchy of modality importance. Removing audio leads to the most substantial performance drop (-1.40\%), particularly in reasoning-intensive tasks (SSR: -2.44\%), highlighting the necessity of auditory cues for social interaction understanding and reasoning. For linguistic cues, the impact is notably asymmetric: while adding subtitles provides only marginal gains (+0.15\%), their removal significantly hinders performance (-0.90\%), especially in SSR and SDP. This suggests that while models may not always benefit from redundant text, the absence of critical textual context creates a significant bottleneck for complex social reasoning. 

\begin{figure}[ht]
    \centering
    \includegraphics[width=0.45\textwidth]{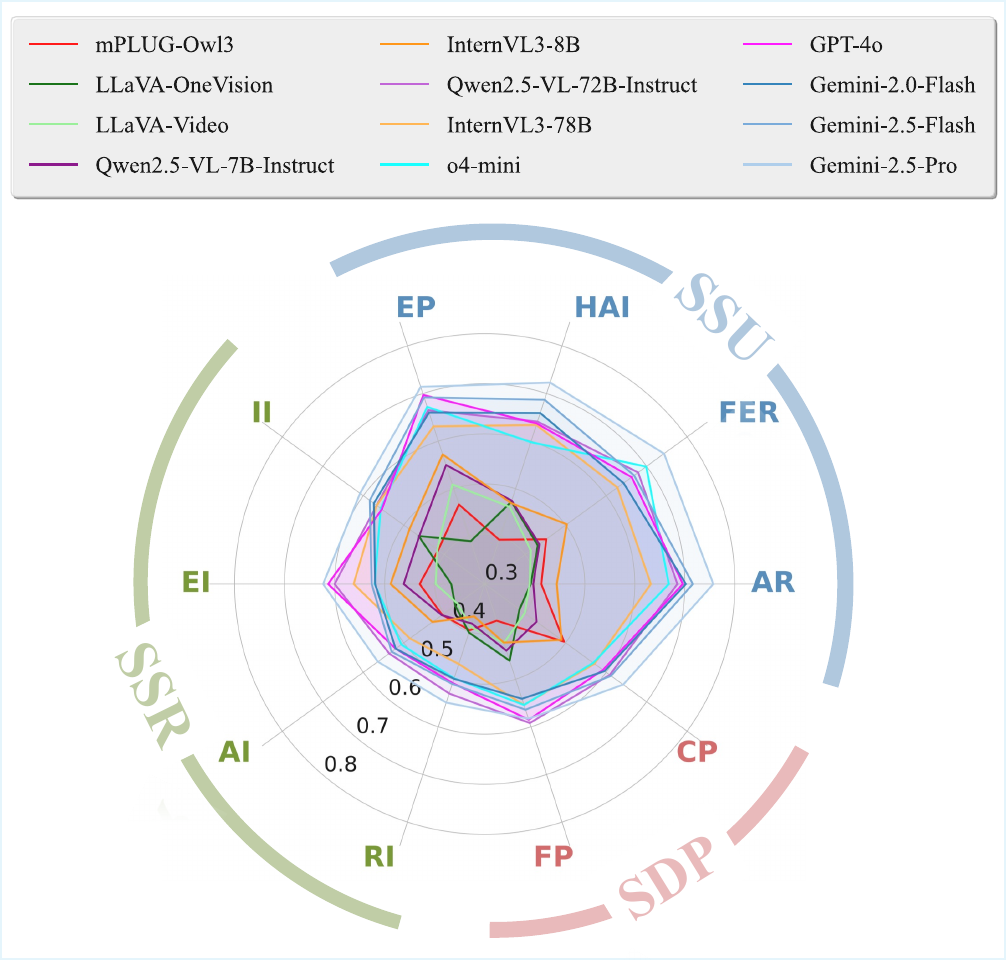}
    \caption{Radar chart of MLLM performance across the 10 fine-grained SIV-Bench sub-tasks.}
    \label{fig:radar}
\end{figure}

\textbf{Task Influence.} SSU focuses on recognizing visible elements and is generally the easiest dimension, where models achieve their highest scores. This is not always the case, since LLaVA-OneVision sometimes performs better on SSR or SDP. In SSU, larger models, including closed-source models and high-parameter open-source models, consistently outperform smaller ones, which likely reflects stronger perceptual ability from greater capacity and broader training \citep{alabdulmohsin2022revisiting}. After SSU, SDP is usually the next most tractable, while SSR remains the most difficult because it requires inferring latent social states such as emotions and intentions. Figure \ref{fig:radar} further breaks down performance across 10 fine-grained tasks in original videos. Stronger closed-source models such as the Gemini and GPT-4o form the outer ring and outperform smaller open-source models. In SSU, the performance gap is most pronounced in Action Recognition (AR) and Facial Expression Recognition (FER), suggesting advantages in capturing subtle visual cues. SSR is more challenging. Models perform moderately on Intent Inference (II) and Emotion Inference (EI), but they struggle most with Relation Inference (RI), which is often the lowest point on the radar. In SDP, performance improves on Factual Prediction (FP) and Counterfactual Prediction (CP), which may reflect social commonsense acquired from language data. Most models perform better on CP than FP, possibly because hypothetical framing provides clearer reasoning cues. Representative failures are shown in Figure \ref{fig:fail ssu}, \ref{fig:fail ssr}, and \ref{fig:fail sdp}.

\subsection{In-Depth Analysis of Relation Inference}
\begin{figure}[ht]
    \centering
    \includegraphics[width=0.48\textwidth]{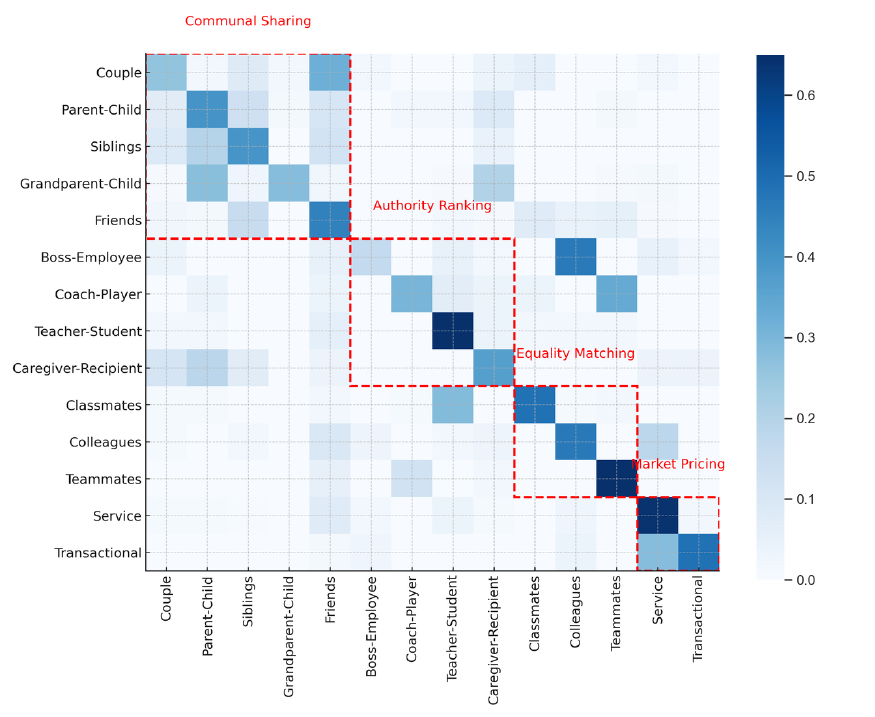}
    \caption{Aggregated Confusion Matrix for the Relation Inference (RI) task across all models. Red dashed lines delineate the four foundational relational models.}
    \label{fig:confusion}
\end{figure}

Since RI emerges as a key bottleneck and provides the relational grounding needed to interpret many higher-level social states, we take a closer look at RI and characterize its error structure quantitatively. Figure \ref{fig:confusion} aggregates RI predictions across the evaluated models into a confusion matrix, which reveals structured error clusters rather than random noise: a prominent pattern is Authority–Equality confusion, where hierarchical relations such as Boss–Employee and Coach–Player are frequently predicted as their egalitarian counterparts (Colleagues and Teammates). Errors also concentrate within relation families. The red dashed boundaries delineate the four foundational relational models, and off-diagonal clusters around these boundaries highlight systematic confusions between conceptually adjacent relation groups. 

Qualitatively, we observe four common failure modes. \textbf{(1)} Models fail to differentiate primary vs. secondary relations in multi-relational scenarios, predicting a plausible but less salient relation. \textbf{(2)} Models are misled by scene- and human-induced cues, over-relying on stereotypical settings or surface language without validating the actual interaction (e.g., classroom \(\rightarrow\) teacher–student). \textbf{(3)} Some errors reflect deficient commonsense social reasoning, where predictions contradict basic social conventions (e.g., two <friends> wearing different sports team uniforms but showing friendly behavior are misclassified as <teammates>, disregarding the fact that such uniforms typically signify competition). \textbf{(4)} Models ignore key perceptual details (e.g., age/role cues) that are decisive for disambiguating closely related relations. 
More failure cases are provided in Appendix \ref{appdx: fail case}.

\subsection{In-depth Analysis of Reasoning Process}
To diagnose the cognitive boundaries of current MLLMs, we establish SIV-Bench-Hard, a subset of 200 QAs where models collectively demonstrate the highest error rates. This focused selection enables concentrated failure-mode analysis and facilitates a reliable human baseline, which is prohibitively expensive at full scale. We recruit three independent crowdworkers, distinct from the original annotators, to provide answer selections and free-text explanations. Table \ref{tab:hard} highlights a substantial performance gap, with Gemini-3-Pro and GPT-5.1 achieving accuracies of 45.50\% and 39.00\%, compared to a human baseline of 74.40\%.


\begin{table}[t!]
    \centering
    \caption{Accuracy and LLM-judge scores (1-5) on reasoning quality (Rel=Relevance, Cohe=Logical Coherence, Depth=Depth of Analysis, Align=Alignment with Human, Conc=Conciseness, Ovrl=Overall).}
    \label{tab:hard}
    \resizebox{0.48\textwidth}{!}{%
    \begin{tabular}{lcccccccc}
        \toprule
        \textbf{Model} & \textbf{Acc\%} & \textbf{Rel} & \textbf{Align} & \textbf{Cohe} & \textbf{Depth} & \textbf{Conc} & \textbf{Ovrl} \\
        \midrule
        Human & 74.40 & - & - & -& -& -& - \\
        \midrule
        Gemini-3-Pro & \textbf{45.50} & \textbf{4.66} & \textbf{3.30}& \textbf{4.67} &	\textbf{3.49} & 4.87&\textbf{4.10} \\
        GPT-5.1 & 39.00 & 4.58 & 3.29 & 4.65 & 3.26 & 4.88 & 4.00 \\
        Gemini-2.5-Pro & 37.00 & 4.57 & 3.26 & 4.65 & 3.41 & 4.89 & 4.05 \\
        Gemini-2.5-Flash & 32.32& 4.48 & 3.17 & 4.55 & 3.22 & 4.87 & 3.95 \\
        GPT-4o-mini & 29.00 & 4.45 & 3.20 & 4.56 & 3.12 & \textbf{4.91} & 3.90 \\
        Qwen2.5-VL-7B & 24.50 & 4.00 & 2.89 & 4.21 & 3.05 & 4.45 & 3.63 \\
        \bottomrule
    \end{tabular}%
    }
\end{table}

LLM-judge results show a gap between reasoning presentation and social cognition. Models achieve strong structural scores (Logical Coherence, Relevance >4.0) but substantially lower Alignment and Depth (<3.5). We also find that reasoning similarity to human traces is positively associated with answer correctness (Figure \ref{fig:reasoning_similarity}, Table \ref{tab:reasoning_similarity_stats}), proving that robust social interaction understanding requires human-like inference rather than purely fluent justifications. Interestingly, while Gemini-3-Pro maintains a lead in analysis depth over GPT-5.1 (+0.23), this does not improve alignment, indicating that an increased depth of reasoning does not inherently guarantee social accuracy.

A primary failure mode involves the inability to identify core social elements. In 40\% to 54\% of failures, models prioritize secondary visual cues over the "long-term separation" or "familial bonds" emphasized by humans. Furthermore, 45\% of depth-related failures stem from a lack of social dynamic analysis, where models stop at direct causality and ignore power structures or cultural norms. We also observe a unique "excessive caution" bias in GPT-5.1, which frequently refuses to make valid socio-cultural inferences from appearance, resulting in the lowest alignment scores where humans readily use such cues for grounding. These insights suggest that MLLMs require multi-layered social grounding that explicitly partitions inference into observational, affective, and normative layers, enabling the models to bridge the gap between literal perception and human-aligned social intelligence.


\section{Related Works}

\textbf{Video Benchmarks for MLLMs.} Rapid progress in video MLLMs has motivated a growing suite of benchmarks spanning perception to high-level reasoning. Representative benchmarks cover multiple tasks (e.g., QA and summarization) over diverse video sources \citep{zhou2024mlvu, fu2024video, li2024mvbench, fang2024mmbench, ning2023video, wu2024chartinsightsevaluatingmultimodallarge}. A major line of work emphasizes long-form understanding and temporal reasoning \citep{rawal2024cinepile, mangalam2023egoschema, wang2024lvbench, liu2024tempcompass}. Meanwhile, emerging benchmarks probe new deployment regimes, such as real-time/streaming video reasoning \citep{xun2025rtv} and egocentric intent understanding with gaze cues \citep{peng2025eye}. While these benchmarks provide broad coverage of video comprehension, they are not designed to systematically evaluate the fine-grained understanding and reasoning required for complex multi-person social interactions.

\textbf{Evaluating Social Intelligence in AI Systems.} Beyond general video understanding, a parallel line of work evaluates social intelligence, moving from component-level signals (e.g., actions and affect) toward richer interpersonal reasoning and grounded social cognition. Recent benchmarks provide narrower but deeper probes, including Theory-of-Mind understanding from short films \citep{villa2025moments}, grounded social reasoning traces from interaction videos \citep{mathur2025social, niu2026read}, and nonverbal social reasoning without speech \citep{li2025mimeqa}. Complementary efforts assess emotional and social intelligence in video QA \citep{zhang2025mme} and social intelligence in authentic multi-turn human conversations \citep{huang2025si}, alongside interactive social environments \citep{zhou2023sotopia, wang2024sotopia, kong2024learning}. However, existing benchmarks largely evaluate social understanding in a task-specific or modality-specific manner. SIV-Bench complements them by providing a unified evaluation of social interaction understanding in real-world videos, spanning perception, social state reasoning, and dynamics prediction.

\section{Conclusion}\label{sec: conclusion}
We introduce \textbf{SIV-Bench}, a video benchmark for evaluating MLLMs on real-world social interaction understanding and reasoning grounded in foundational social relationship models. SIV-Bench contains 2,792 videos and 5,455 curated QA pairs, organized into three dimensions (SSU, SSR, and SDP) as well as ten fine-grained sub-tasks for diagnosis. Extensive experiments show that current MLLMs perform relatively well on SSU but struggle with high-level reasoning, with Relation Inference exhibiting systematic confusions across relation types. Subtitle ablations suggest that textual signals can affect performance in a task-dependent manner, particularly for complex inference cases. On a challenging hard subset, we further observe a substantial gap to human performance, reinforcing the difficulty of human-aligned social reasoning. We hope SIV-Bench will serve as a unified testbed to track progress and drive more robust social intelligence in MLLMs.

\section*{Limitations}
While SIV-Bench offers a structured diagnostic benchmark for social interaction understanding, the current size and coverage still leave room to expand toward broader social settings, cultures, and interaction styles. Moreover, our primary evaluation remains multiple-choice, which simplifies scoring but does not fully capture open-ended grounding, generation, or interactive behaviors. We include SIV-Bench-Hard with brief explanations to partially address this gap, yet it is relatively small and still constrained in evaluation scope. Finally, social interpretation can be subjective and culturally contingent, and some clips may admit multiple plausible readings, suggesting the need for future work on more calibrated human protocols and richer evaluation formats.

\section*{Ethical Considerations}
SIV-Bench is built from publicly available videos (e.g., TikTok and YouTube), and we carefully consider licensing and privacy risks when creating and releasing the benchmark. To reduce copyright and platform-policy concerns, we will not redistribute raw video files; instead, we will release annotations along with video identifiers (e.g., URLs and timestamps) and provide a script for users to retrieve content directly from the original platforms, following common practice in prior video benchmarks.

Because the videos may contain identifiable individuals, we applied manual filtering to exclude content that appears overly private or sensitive, and we encourage responsible downstream use. Our human annotation procedures were designed to be minimal-risk, with annotators compensated and handled anonymously. Finally, while SIV-Bench is intended to support transparent evaluation of MLLMs’ social understanding, we acknowledge the dual-use nature of social inference technologies and recommend that users avoid deployment in high-stakes settings and follow applicable privacy and platform guidelines when using the benchmark.

\section*{Acknowledgments}
This work was supported by the Beijing Natural Science Foundation under Grant No. 4264117.

\bibliography{custom}

\appendix

\section{Video Details}
\subsection{Video Collection}
\begin{figure}[htbp]
    \centering

    \begin{subfigure}[t]{0.48\columnwidth}
        \centering
        \includegraphics[width=\linewidth]{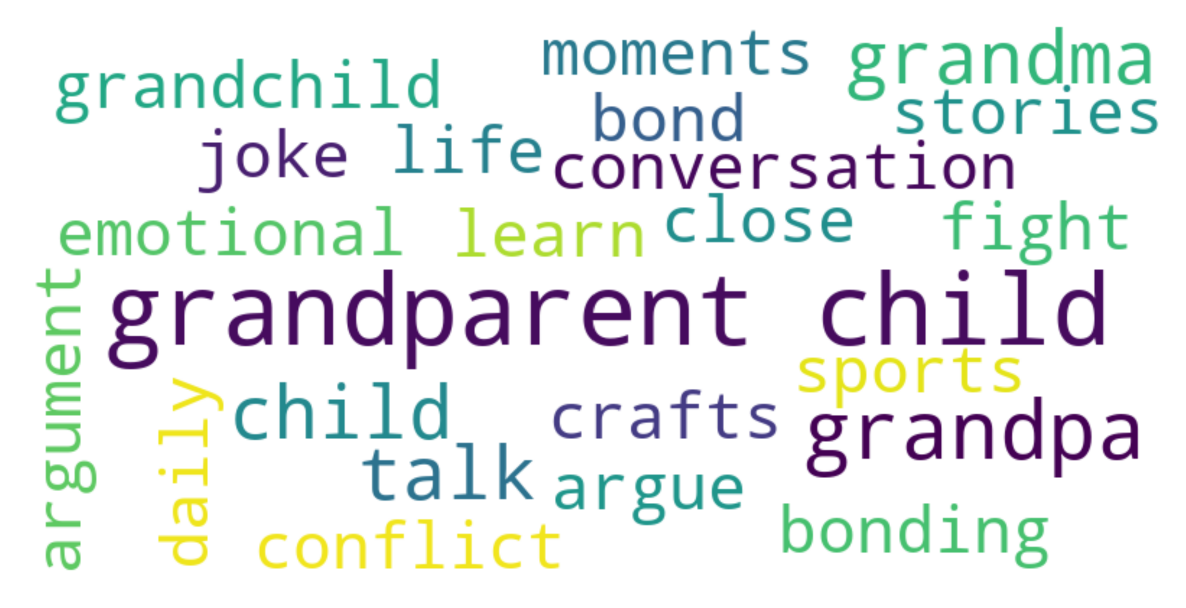}
        \caption{Grandparent--Child}
    \end{subfigure}\hfill
    \begin{subfigure}[t]{0.48\columnwidth}
        \centering
        \includegraphics[width=\linewidth]{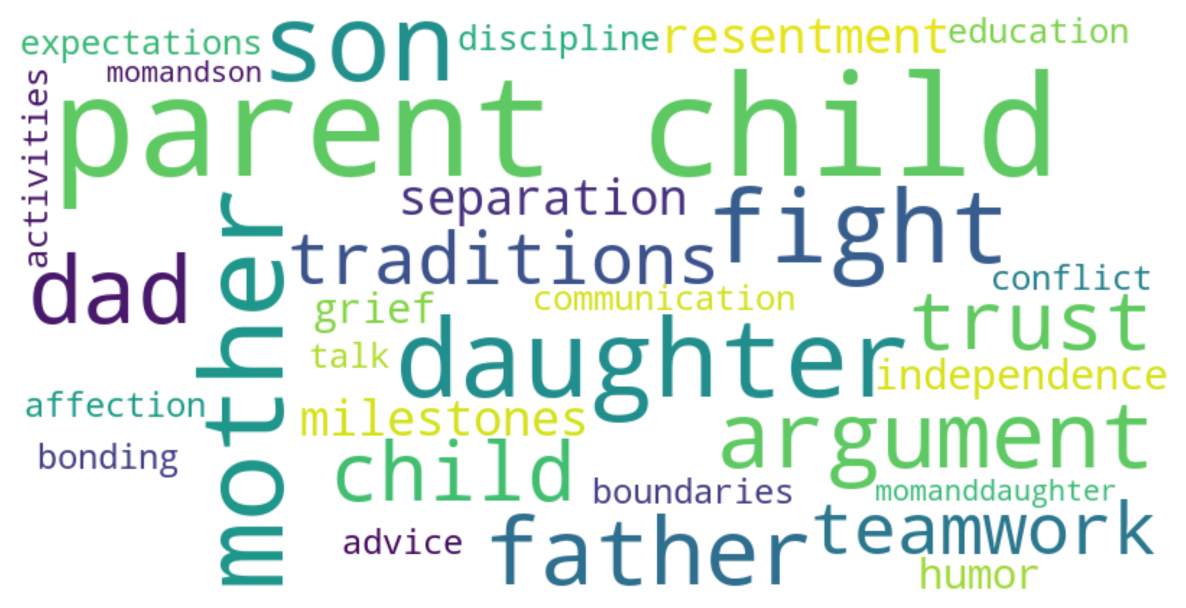}
        \caption{Parent--Child}
    \end{subfigure}

    \medskip

    \begin{subfigure}[t]{0.48\columnwidth}
        \centering
        \includegraphics[width=\linewidth]{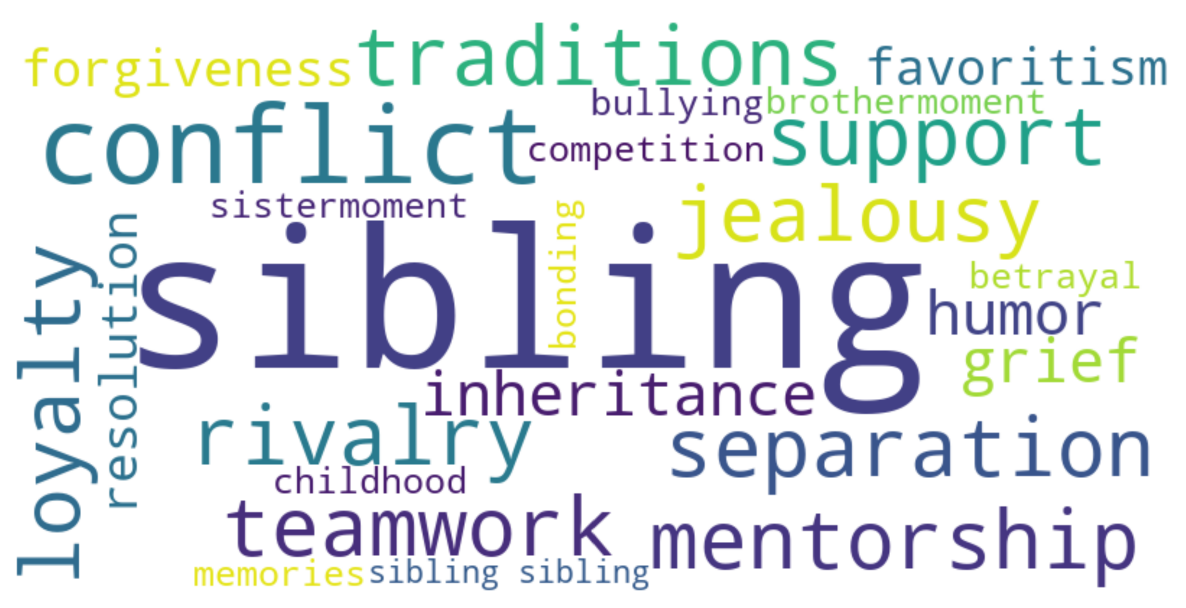}
        \caption{Siblings}
    \end{subfigure}\hfill
    \begin{subfigure}[t]{0.48\columnwidth}
        \centering
        \includegraphics[width=\linewidth]{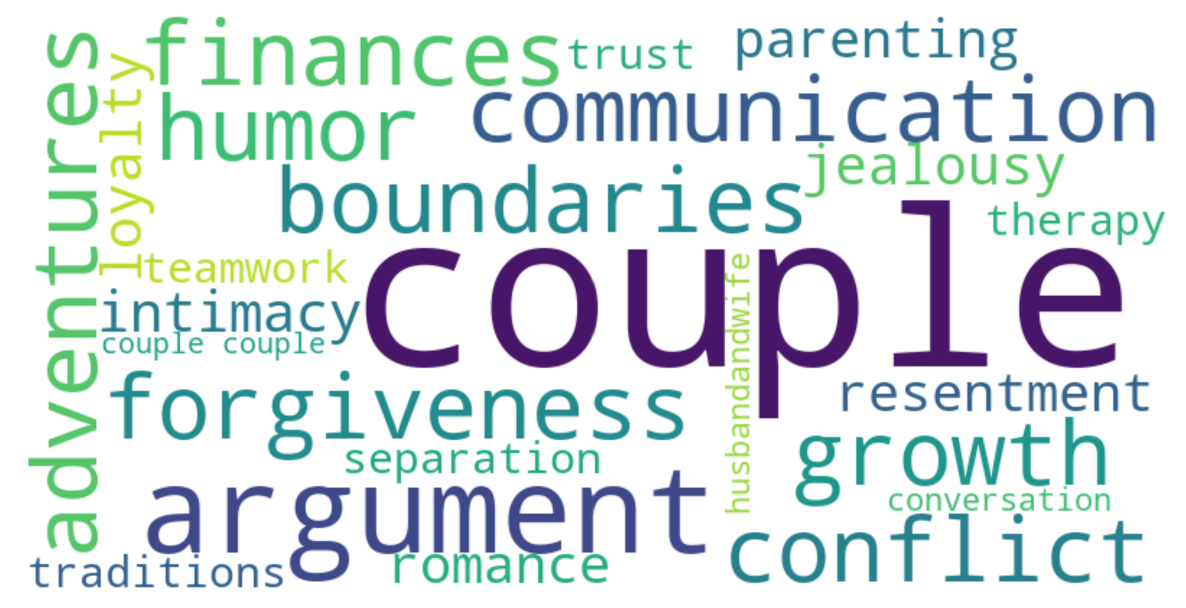}
        \caption{Couple}
    \end{subfigure}

    \medskip

    \begin{subfigure}[t]{0.48\columnwidth}
        \centering
        \includegraphics[width=\linewidth]{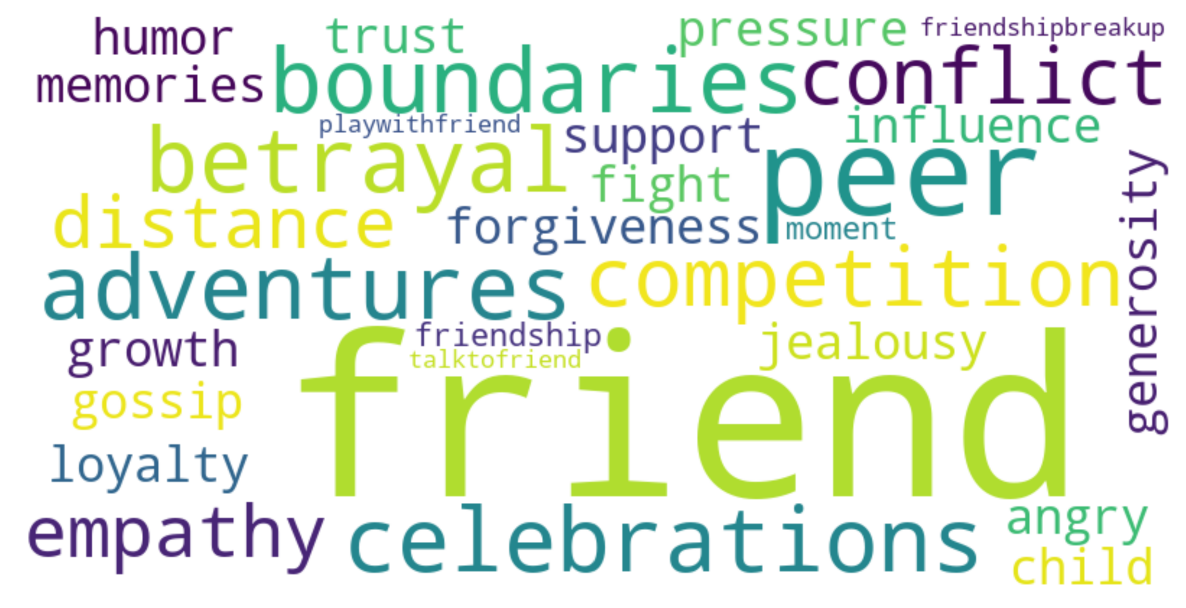}
        \caption{Friends}
    \end{subfigure}\hfill
    \begin{subfigure}[t]{0.48\columnwidth}
        \centering
        \includegraphics[width=\linewidth]{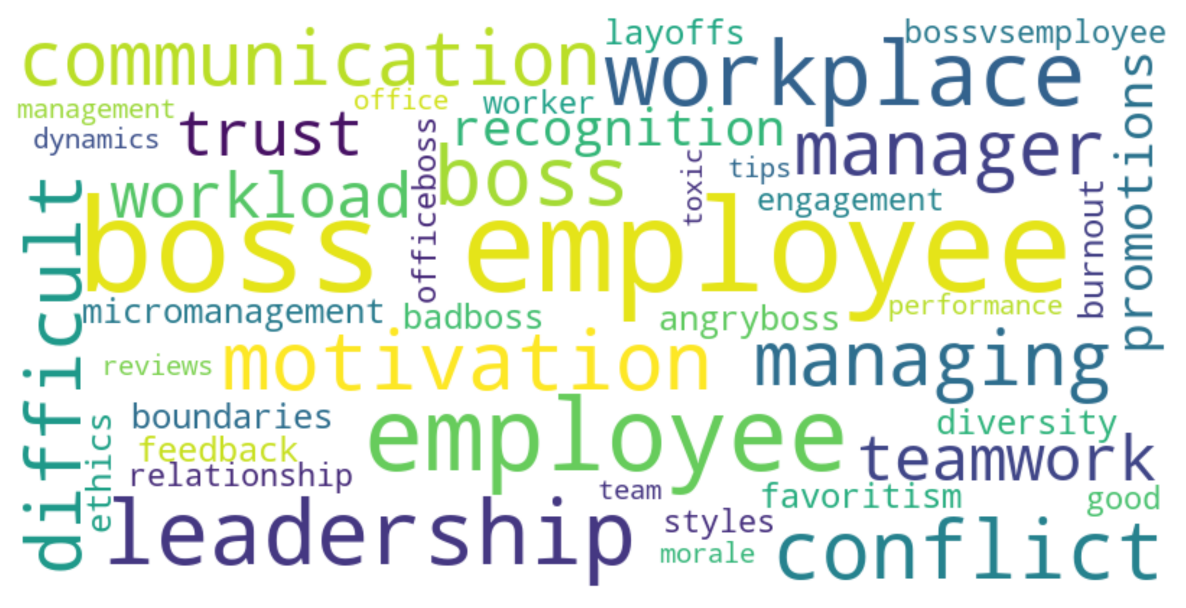}
        \caption{Boss--Employee}
    \end{subfigure}

    \medskip

    \begin{subfigure}[t]{0.48\columnwidth}
        \centering
        \includegraphics[width=\linewidth]{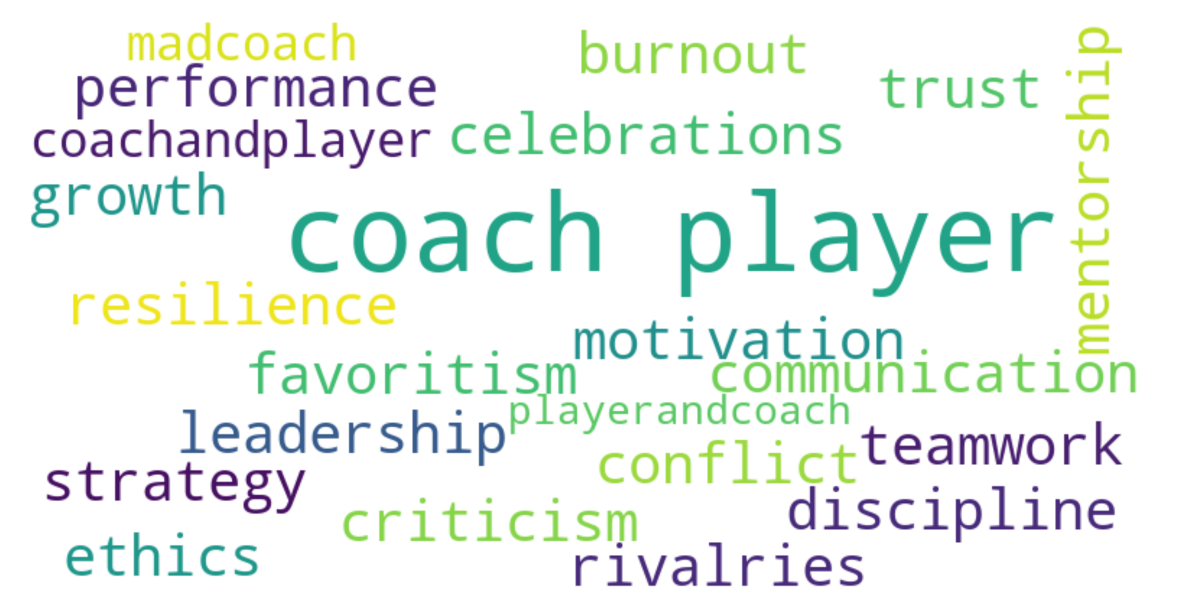}
        \caption{Coach--Player}
    \end{subfigure}\hfill
    \begin{subfigure}[t]{0.48\columnwidth}
        \centering
        \includegraphics[width=\linewidth]{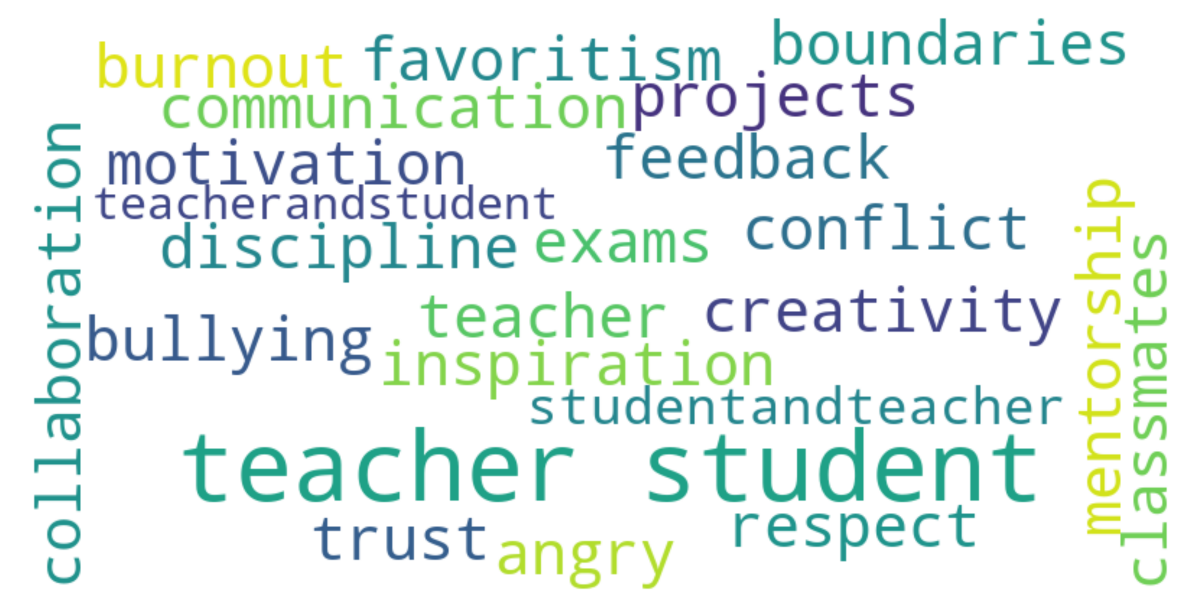}
        \caption{Teacher--Student}
    \end{subfigure}

    \medskip

    \begin{subfigure}[t]{0.48\columnwidth}
        \centering
        \includegraphics[width=\linewidth]{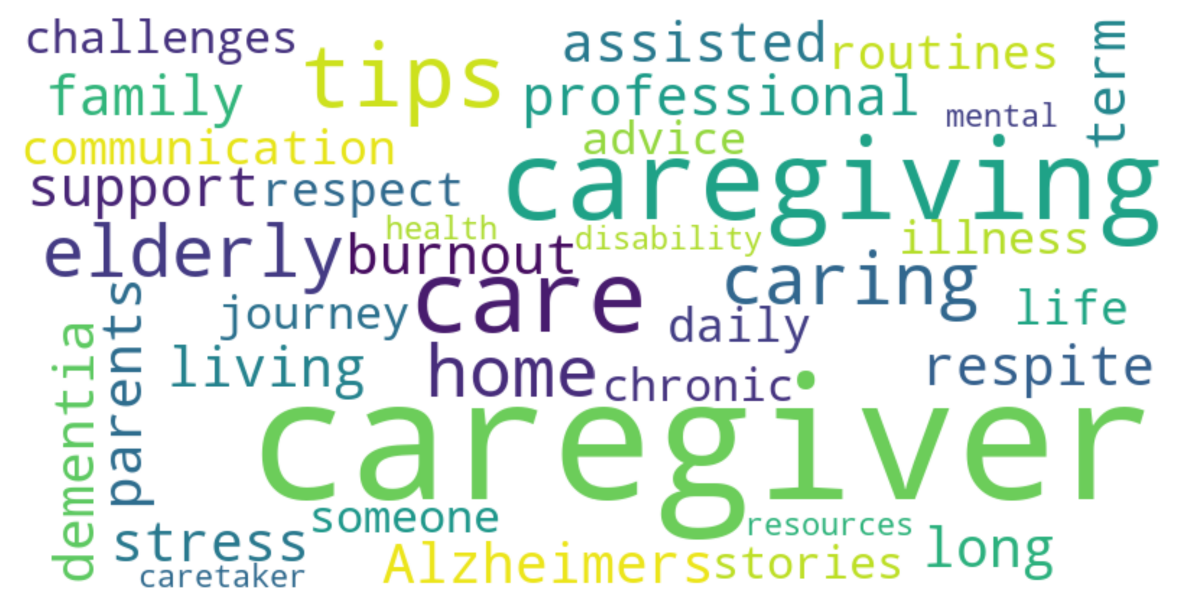}
        \caption{Caregiver--Recipient}
    \end{subfigure}\hfill
    \begin{subfigure}[t]{0.48\columnwidth}
        \centering
        \includegraphics[width=\linewidth]{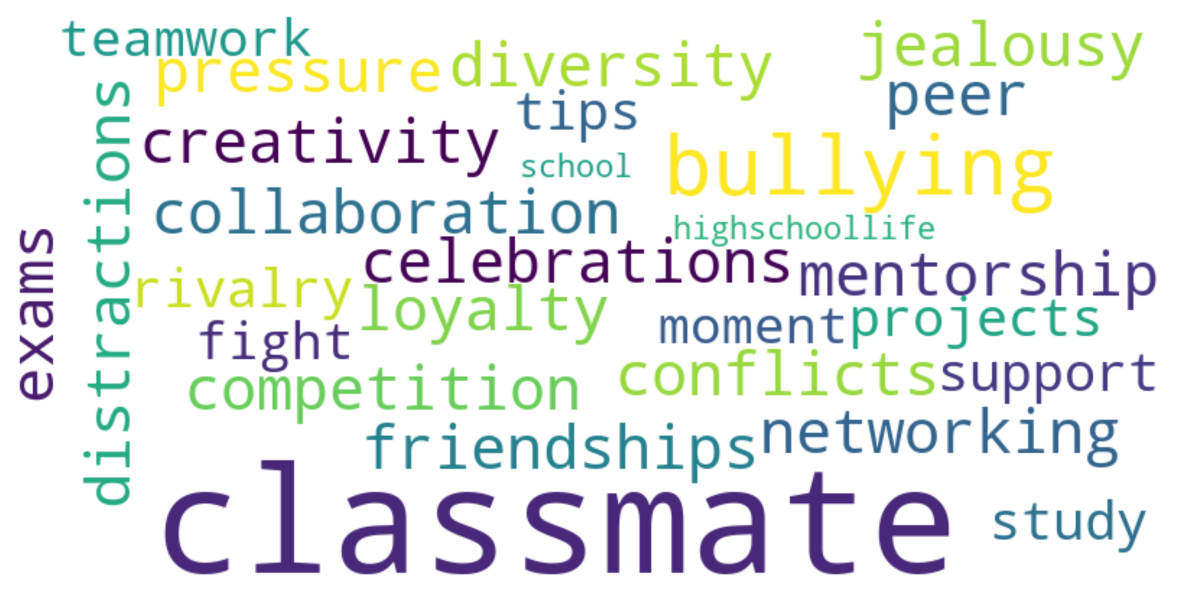}
        \caption{Classmates}
    \end{subfigure}

    \medskip

    \begin{subfigure}[t]{0.48\columnwidth}
        \centering
        \includegraphics[width=\linewidth]{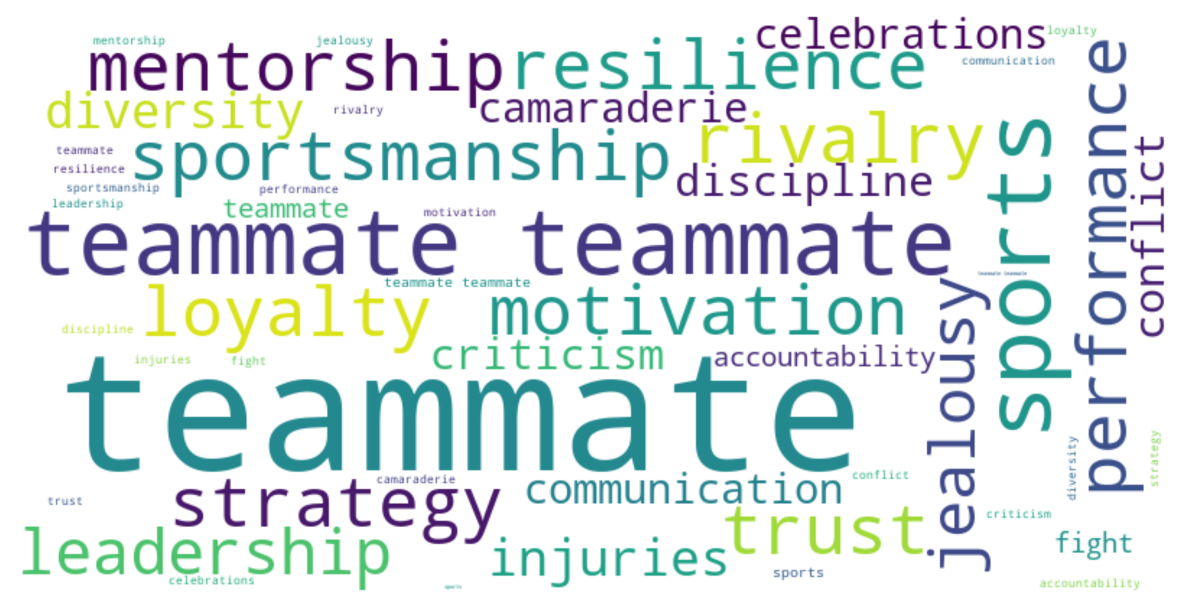}
        \caption{Teammates}
    \end{subfigure}\hfill
    \begin{subfigure}[t]{0.48\columnwidth}
        \centering
        \includegraphics[width=\linewidth]{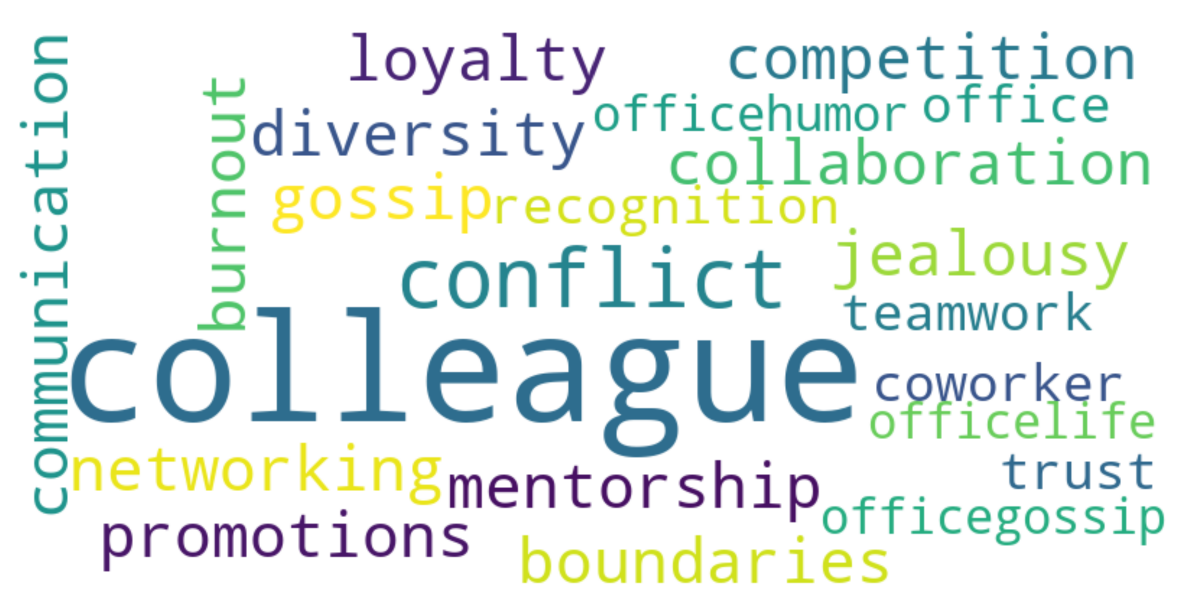}
        \caption{Colleague}
    \end{subfigure}

    \medskip

    \begin{subfigure}[t]{0.48\columnwidth}
        \centering
        \includegraphics[width=\linewidth]{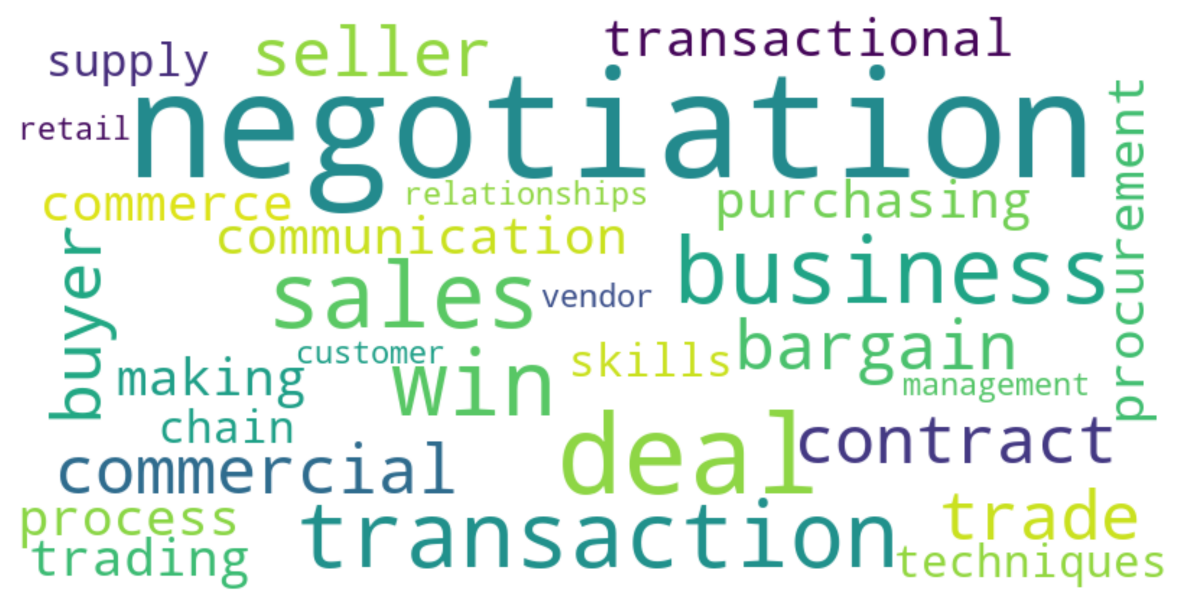}
        \caption{Transactional}
    \end{subfigure}\hfill
    \begin{subfigure}[t]{0.48\columnwidth}
        \centering
        \includegraphics[width=\linewidth]{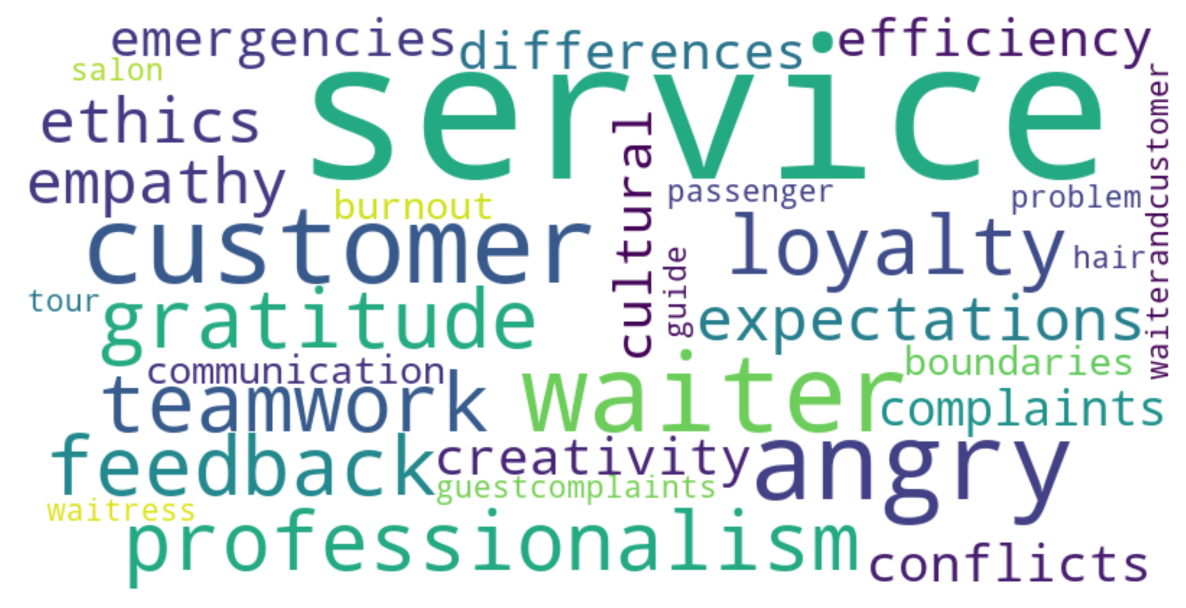}
        \caption{Service}
    \end{subfigure}

    \caption{
    Word clouds of GPT-generated keywords used for sourcing videos across 14 distinct social relationship types.
    Keywords capture diverse interaction scenarios, roles, and intimacy levels from platforms such as YouTube and TikTok.
    }
    \label{fig:keywords_cloud}
\end{figure}

To compile a rich and diverse video corpus for SIV-Bench that reflects a wide array of social interactions, we first identify 14 key social relationship types (e.g., parent-child, friends, colleagues, service interactions).  For each of these types, we employ GPT-4o mini to generate a comprehensive set of search keywords.  This strategy is designed to ensure broad coverage in our video collection from platforms like YouTube and TikTok, capturing various interaction categories (such as cooperation, conflict, and support), forms (including verbal discussions, non-verbal expressions, and shared activities), and differing levels of intimacy or formality inherent to each relationship.  Figure \ref{fig:keywords_cloud} visualizes these tailored keyword sets as word clouds for each of the 14 relationship types, illustrating the breadth of concepts targeted.  This targeted yet diverse keyword generation is crucial for assembling a video dataset that truly represents the complexity of human social life.

\subsection{Video Diversity} \label{appdx: video diversity}
\begin{figure}[htbp]
    \centering
    \includegraphics[width=0.98\linewidth]{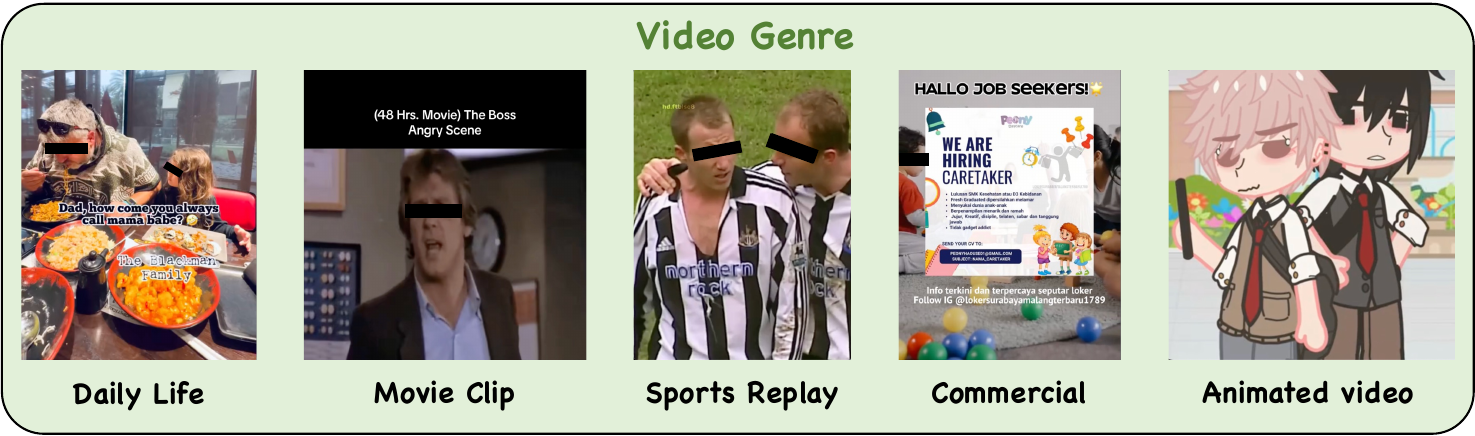}
    \caption{Examples illustrating the diverse video genres in SIV-Bench, including (from left to right) Daily Life recordings, Movie Clips, Sports Replays, Commercials, and Animated Videos, ensuring a broad range of social interaction contexts.}
    \label{fig:video genre}
\end{figure}

\begin{figure}[ht]
    \centering
    \includegraphics[width=0.98\linewidth]{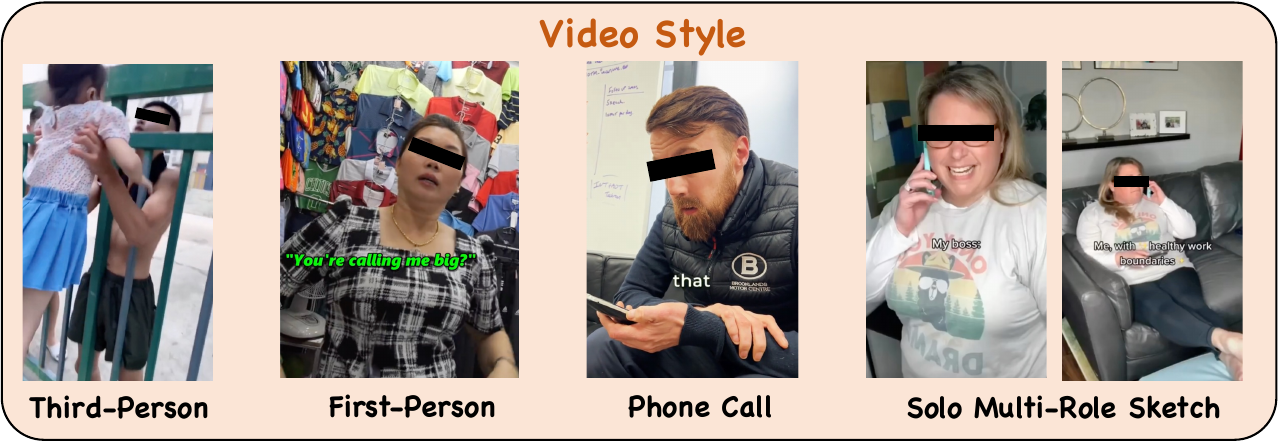}
    \caption{Examples of diverse video presentation styles featured in SIV-Bench, including (from left to right) Third-Person views, First-Person perspectives, Phone Call interfaces, and Solo Multi-Role Sketches.}
    \label{fig:video style}
\end{figure}

The SIV-Bench video corpus is intentionally diverse in both subject matter and presentation to ensure comprehensive model evaluation. This genre diversity, illustrated in Figure \ref{fig:video genre}, includes a wide array of video types such as candid Daily Life recordings, scripted Movie Clips, dynamic Sports Replays, illustrative Commercials, and non-photorealistic Animated Videos. Complementing this, Figure \ref{fig:video style} showcases varied visual presentation styles, ranging from traditional Third-Person perspectives and immersive First-Person (POV) footage to interactions depicted via Phone Call interfaces and creative Solo Multi-Role Sketches. This multifaceted diversity in video content and form provides a robust and varied testbed.
\begin{figure}[htbp]
    \centering
    \includegraphics[width=0.8\linewidth]{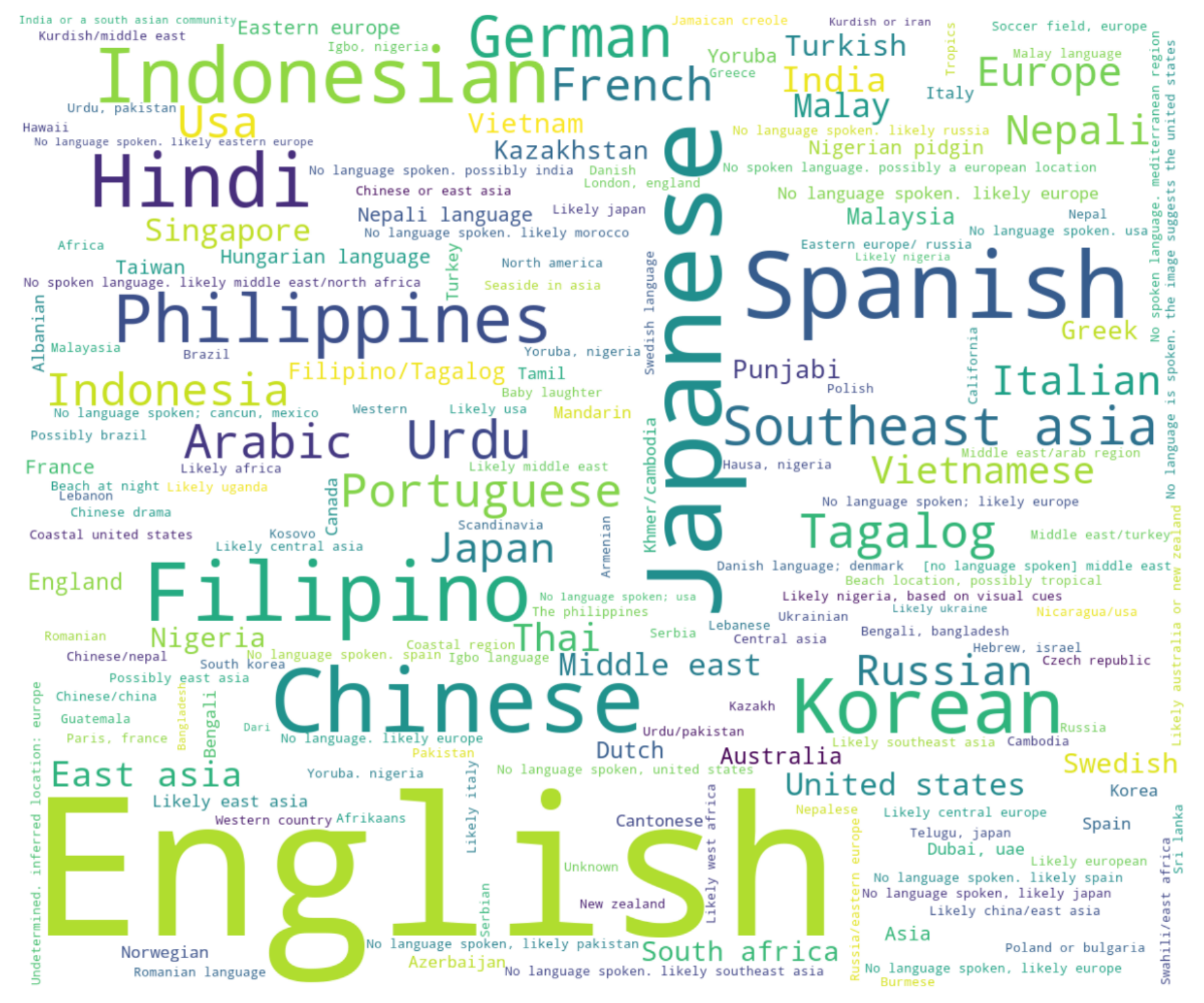}
    \caption{Language word-cloud in SIV-Bench.}
    \label{fig: language}
\end{figure}

\section{QA Details}
\subsection{Automatically QA Generation} \label{appdx: qa generation}

\begin{figure*}[ht]
    \centering
    \includegraphics[width=0.99\linewidth]{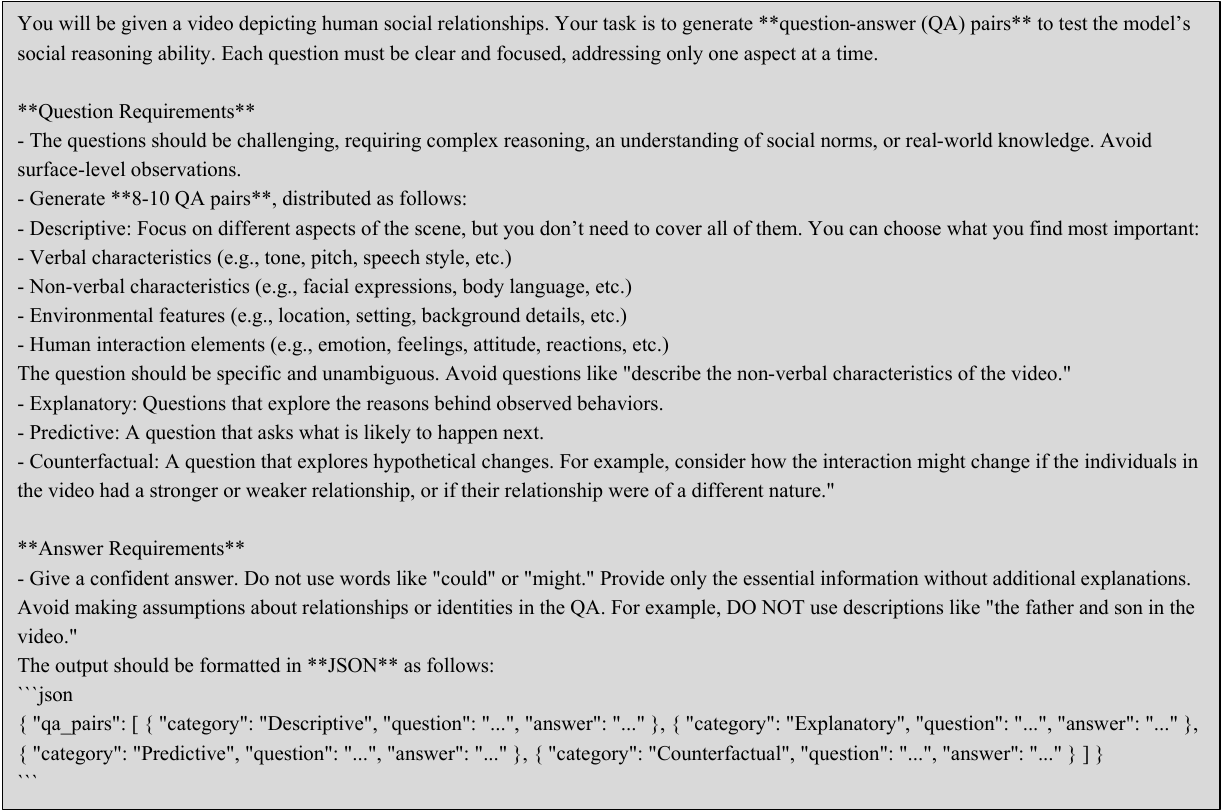}
    \caption{The prompt used to guide Gemini for the initial generation of question-answer pairs.}
    \label{fig:qa_generation_prompt}
\end{figure*}

The prompt shown in Figure \ref{fig:qa_generation_prompt} guides the Gemini in generating diverse and challenging Question-Answer (QA) pairs that form the foundation of SIV-Bench. Its design ensures these QAs effectively cover our core evaluation dimensions.

\begin{figure*}[ht]
    \centering
    \includegraphics[width=0.99\linewidth]{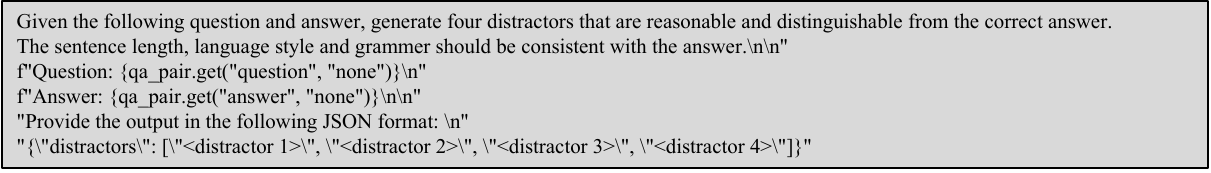}
    \caption{The prompt used to generate distractors for each QA pair.}
    \label{fig:prompt distractors}
\end{figure*}

To create the multiple-choice options for each QA pair, we utilize a dedicated prompt for distractor generation, as detailed in Figure \ref{fig:prompt distractors}. This prompt instructs GPT to produce four unique distractor options based on the given question and its corresponding correct answer. Crucially, the prompt emphasizes that these distractors must be reasonable and clearly distinguishable from the correct answer, while also maintaining consistency with the correct answer in terms of sentence length, language style, and grammar. Furthermore, the prompt specifies a JSON output format for the list of distractors to ensure seamless programmatic integration. This structured approach to distractor generation is essential for constructing high-quality, fair multiple-choice questions that minimize biases arising from option formatting and robustly test model comprehension.

\begin{figure*}[ht]
    \centering
    \includegraphics[width=0.99\linewidth]{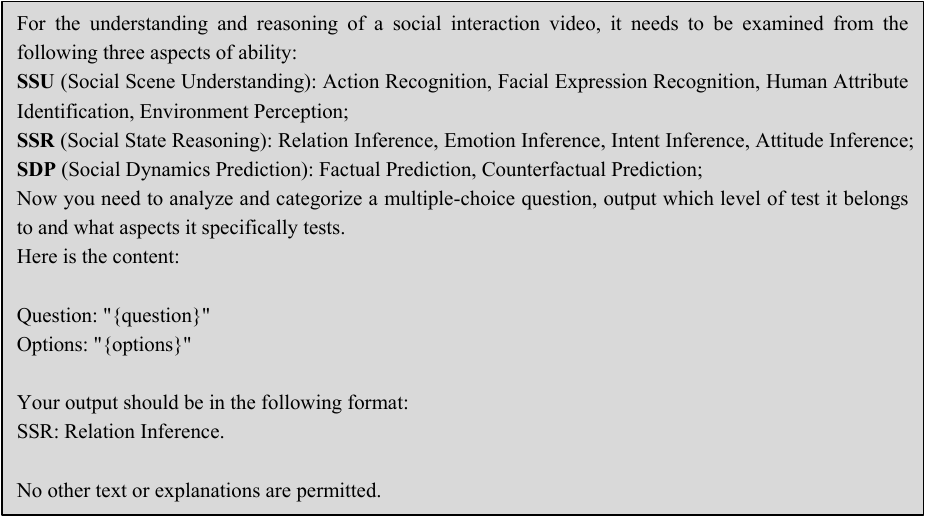}
    \caption{Prompt used to instruct LLM for the final classification of Question-Answer pairs into one of the 10 fine-grained sub-tasks under SSU, SSR and SDP.}
    \label{fig:prompt-task cls}
\end{figure*}

To classify each QA pair in SIV-Bench, we use a GPT model guided by a structured prompt (see Figure \ref{fig:prompt-task cls}). The prompt defines our three main assessment dimensions—SSU, SSR, and SDP—and their sub-tasks. The model analyzes each QA and outputs its corresponding sub-task, ensuring consistent and precise categorization across the benchmark.

To classify each QA pair in SIV-Bench, we use a GPT model guided by a structured prompt (see Figure \ref{fig:prompt-task cls}). The prompt defines our three main assessment dimensions—SSU, SSR, and SDP—and their sub-tasks. The model analyzes each QA and outputs its corresponding sub-task, ensuring consistent and precise categorization across the benchmark.

\subsection{Human Annotation} \label{appdx: human}
\textbf{Annotator Information.} For the human annotation tasks integral to SIV-Bench, we recruit a team of 20 annotators. This cohort consists of 12 female and 8 male individuals, with an average age of 27 years. All annotators are well-educated and demonstrate proficiency in English, which is the primary language for instructions and annotations. To ensure fair compensation for their detailed work, annotators are remunerated at an hourly rate of \$4 USD.

\begin{figure}[h!]
    \centering
    \includegraphics[width=0.98\linewidth]{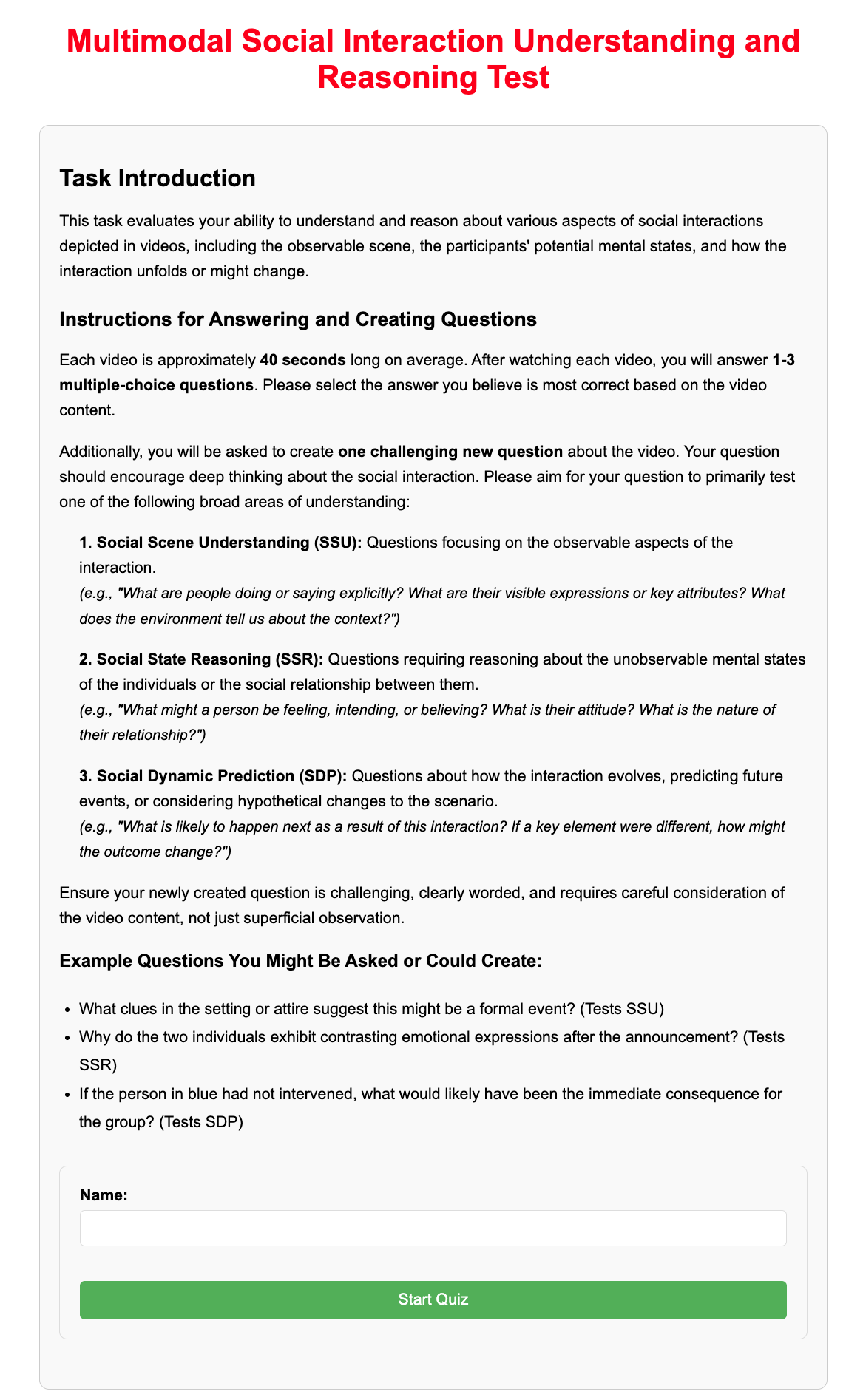}
    \caption{Screenshot of the guidelines provided to human annotators, detailing the tasks of answering multiple-choice questions and creating new challenging questions about Social Scene Understanding (SSU), Social State Reasoning (SSR), and Social Dynamics Prediction (SDP).}
    \label{fig:human inst}
\end{figure}

\textbf{Annotator Guidelines.} To maintain consistency and high quality in human annotation, particularly for the creation of new challenging Question-Answers (QAs) for Subset 2, annotators are provided with a comprehensive set of guidelines, as illustrated in Figure \ref{fig:human inst}. These instructions detail the process for responding to existing multiple-choice questions and, crucially, guide annotators in formulating one new challenging question per video. For this QA creation task, annotators are directed to ensure their questions primarily test one of our three core assessment dimensions: Social Scene Understanding (SSU), Social State Reasoning (SSR), or Social Dynamics Prediction (SDP). The guidelines, including example questions, also emphasize that new QAs must be demanding, clearly worded, and require a deep understanding of the video content rather than superficial observation.

\begin{figure}[h!]
    \centering
    \includegraphics[width=0.95\linewidth]{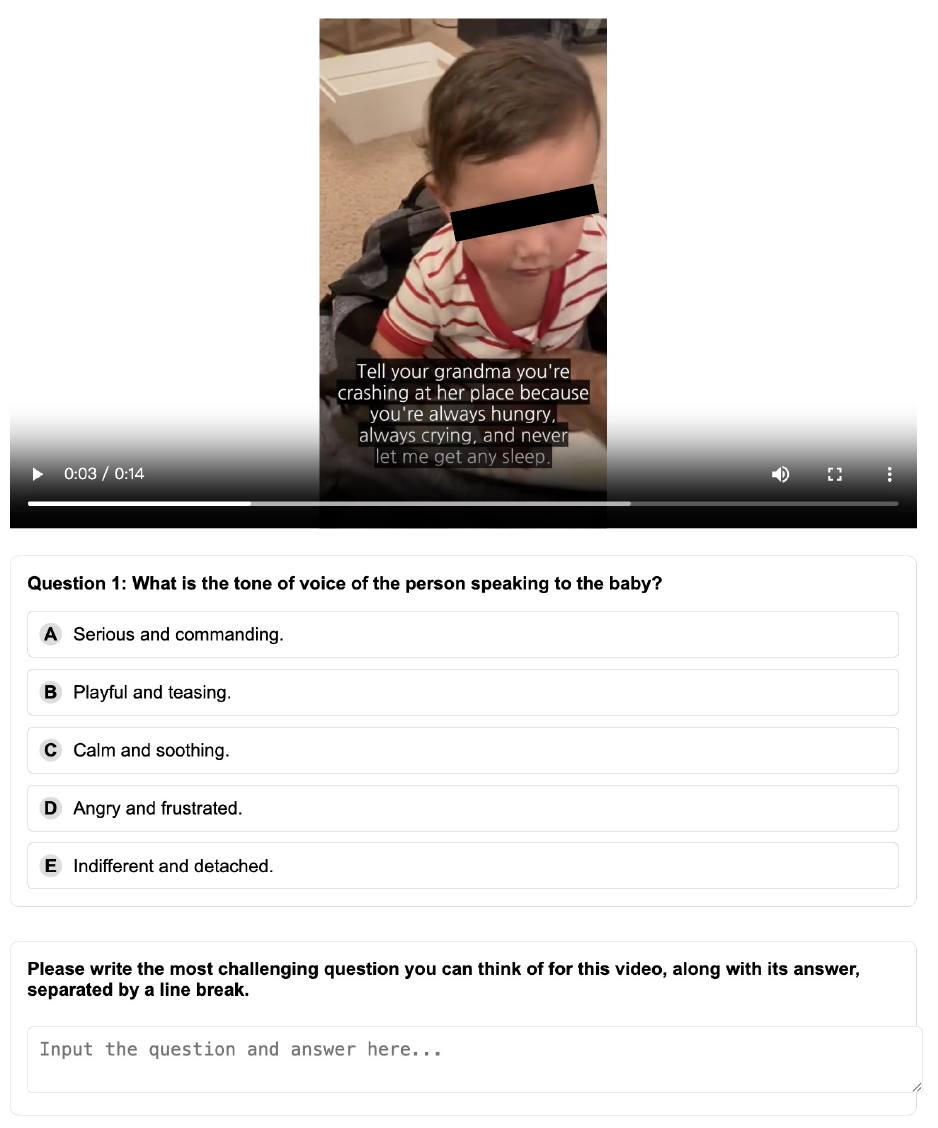}
    \caption{Example of the web-based interface used by human annotators for watching videos, answering provided multiple-choice questions, and authoring new Question-Answer pairs for SIV-Bench.}
    \label{fig:human example}
\end{figure}

\textbf{Annotation Process.} All human annotation tasks for SIV-Bench, including the review of existing Question-Answers (QAs) and the generation of new QAs, are conducted using a custom-designed web interface. Figure \ref{fig:human example} provides a representative example of this interface, which allows annotators to view the social interaction video, respond to provided multiple-choice questions, and submit their own newly authored questions and answers based on the video content and provided guidelines. This platform ensures a standardized environment for all human annotation contributions.

\textbf{Quality Control and Inter-Annotator Agreement.} To quantitatively address label quality, we compute the Inter-Annotator Agreement (IAA) using Fleiss' Kappa. For the initial set of approximately 27,000 QA pairs generated by our three LLM experts, the resulting Fleiss' Kappa is 0.52, indicating "moderate" agreement. For the questions that the LLMs disagreed, we measure the agreement between our human annotators, yielding a Fleiss' Kappa of 0.68. According to established standards \citep{landis1977measurement}, this value represents "substantial" agreement. Achieving this on the most difficult portion of our dataset confirms that humans can establish a reliable ground truth and justifies our methodology of retaining only those questions on which all assigned annotators unanimously agreed for the final benchmark.

\subsection{QA statistics} \label{appdx: qa stats}
\begin{figure*}[t]
    \centering

    \begin{subfigure}[t]{0.48\textwidth}
        \centering
        \includegraphics[width=\linewidth]{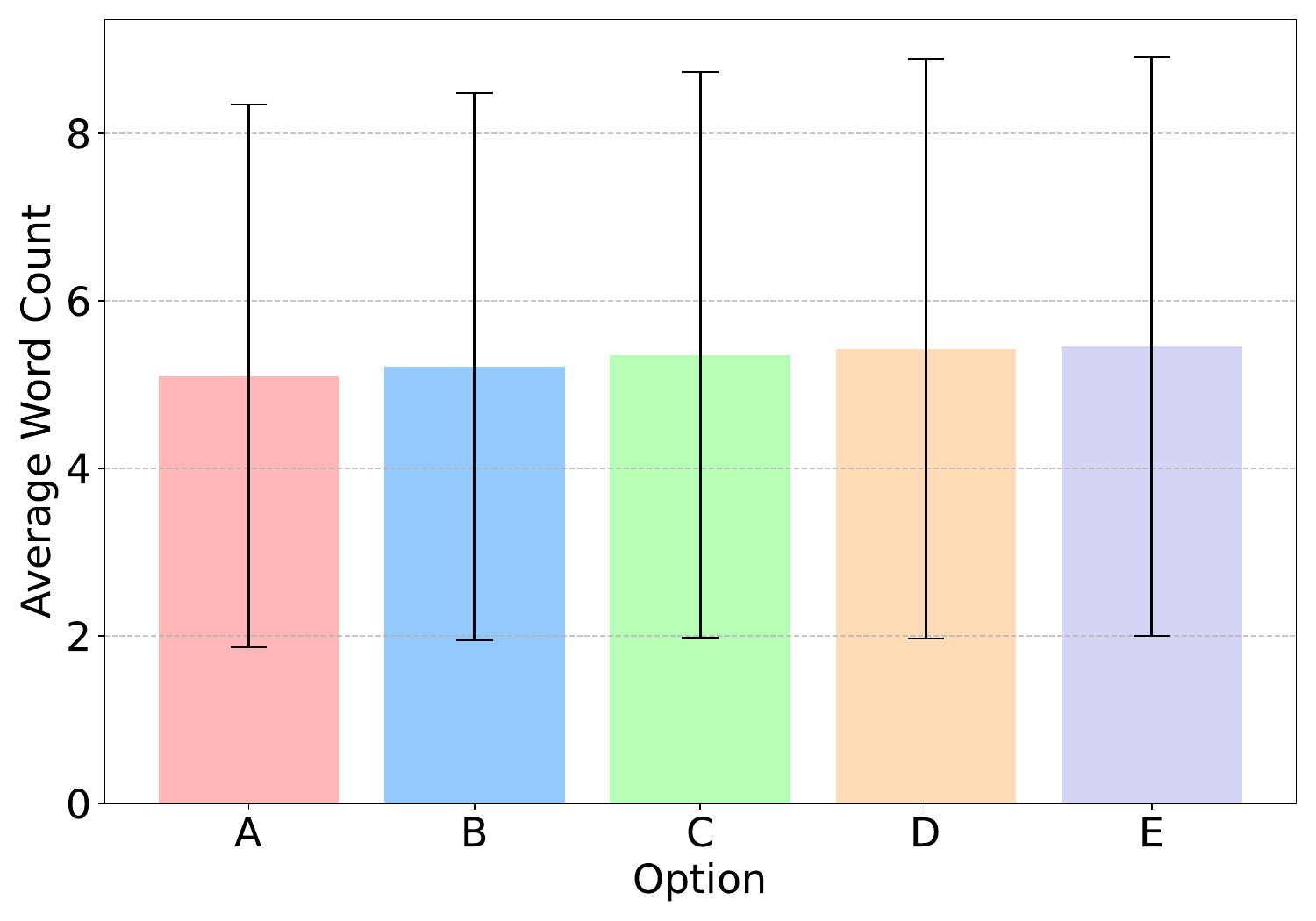}
        \caption{Word count per option}
        \label{fig:average_word_count_per_option}
    \end{subfigure}\hfill
    \begin{subfigure}[t]{0.4\textwidth}
        \centering
        \includegraphics[width=\linewidth]{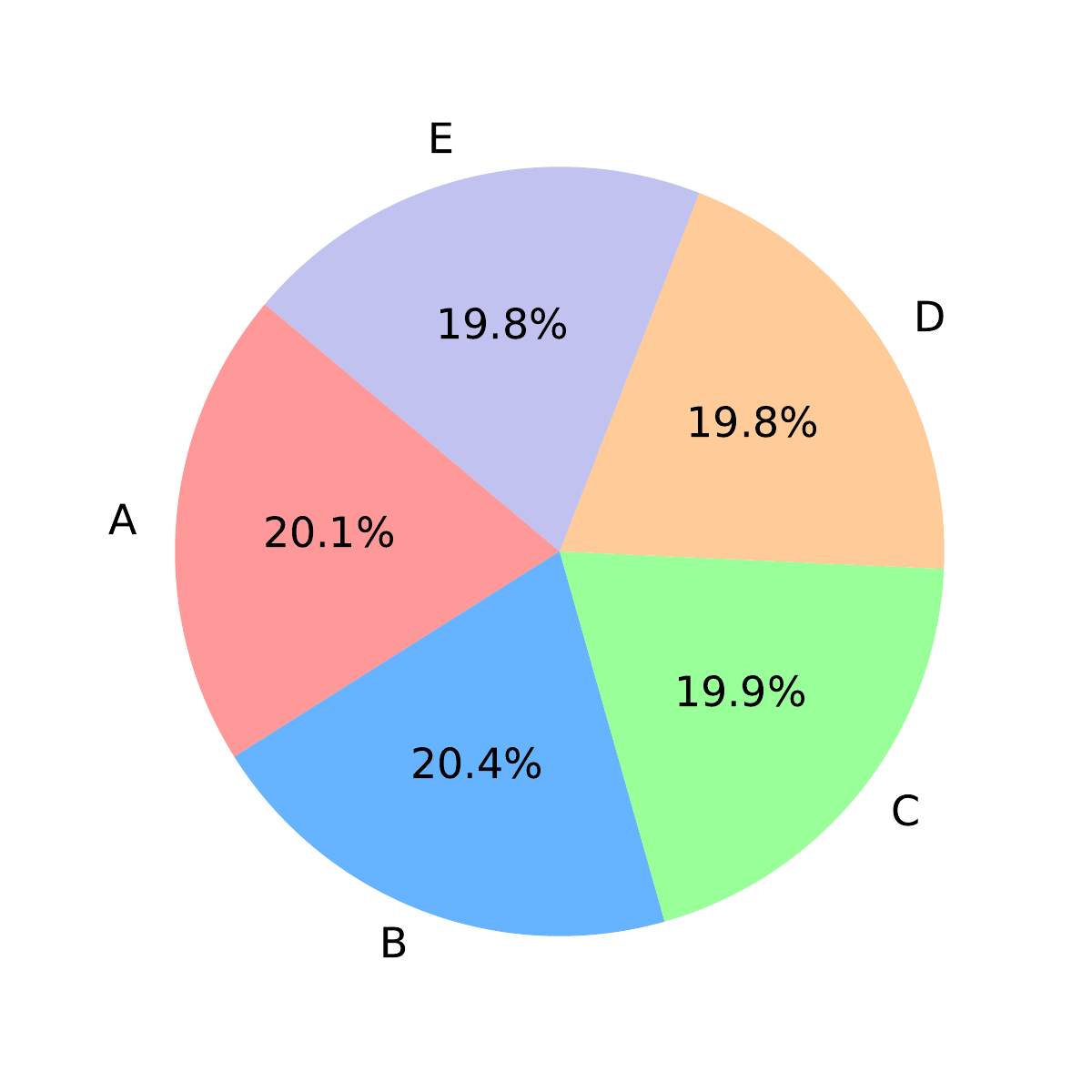}
        \caption{Correct answer distribution}
        \label{fig:correct_answer_distribution}
    \end{subfigure}

    \medskip

    \begin{subfigure}[t]{0.48\textwidth}
        \centering
        \includegraphics[width=\linewidth]{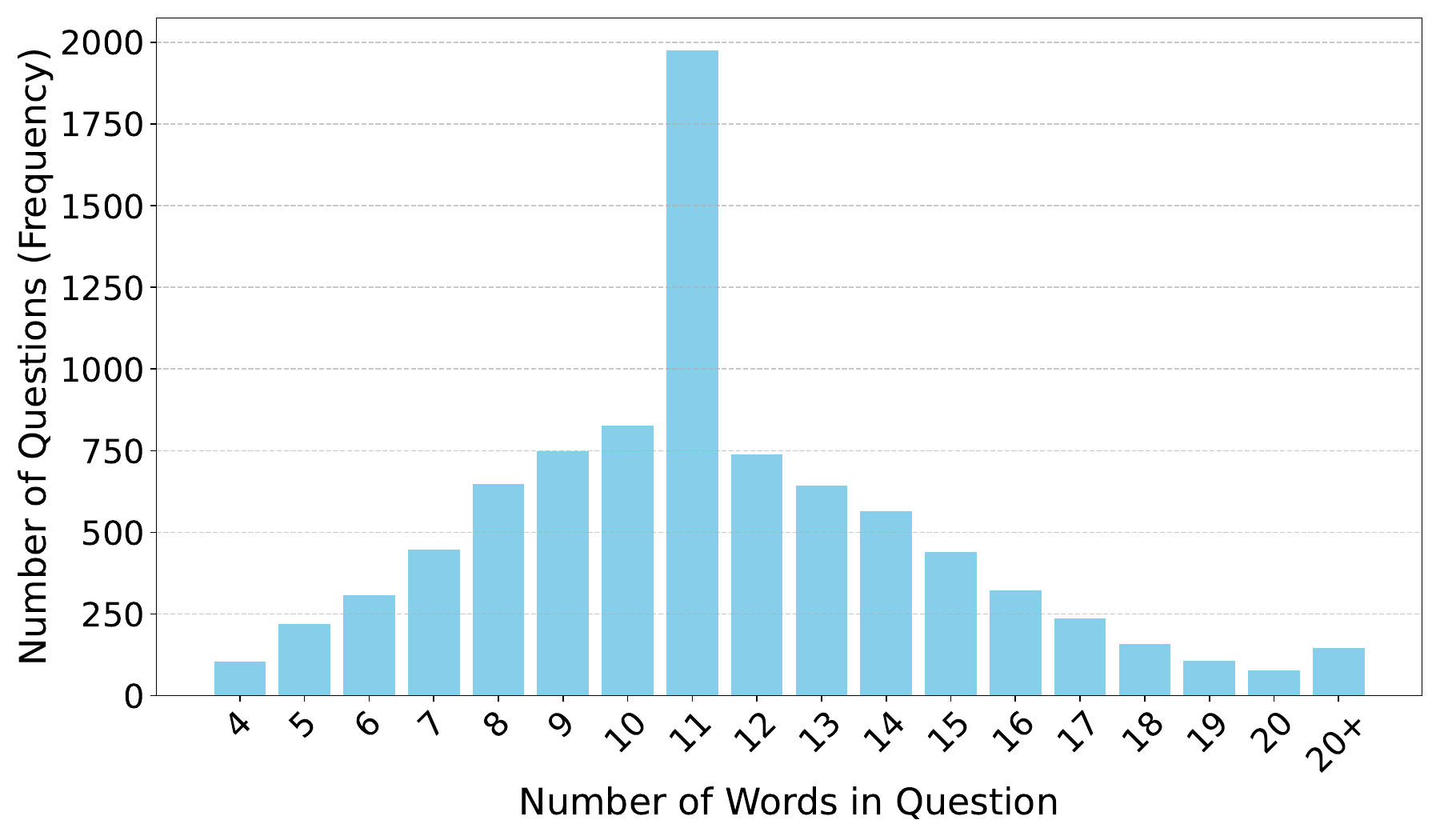}
        \caption{Question word count}
        \label{fig:question_word_count_distribution}
    \end{subfigure}\hfill
    \begin{subfigure}[t]{0.4\textwidth}
        \centering
        \includegraphics[width=\linewidth]{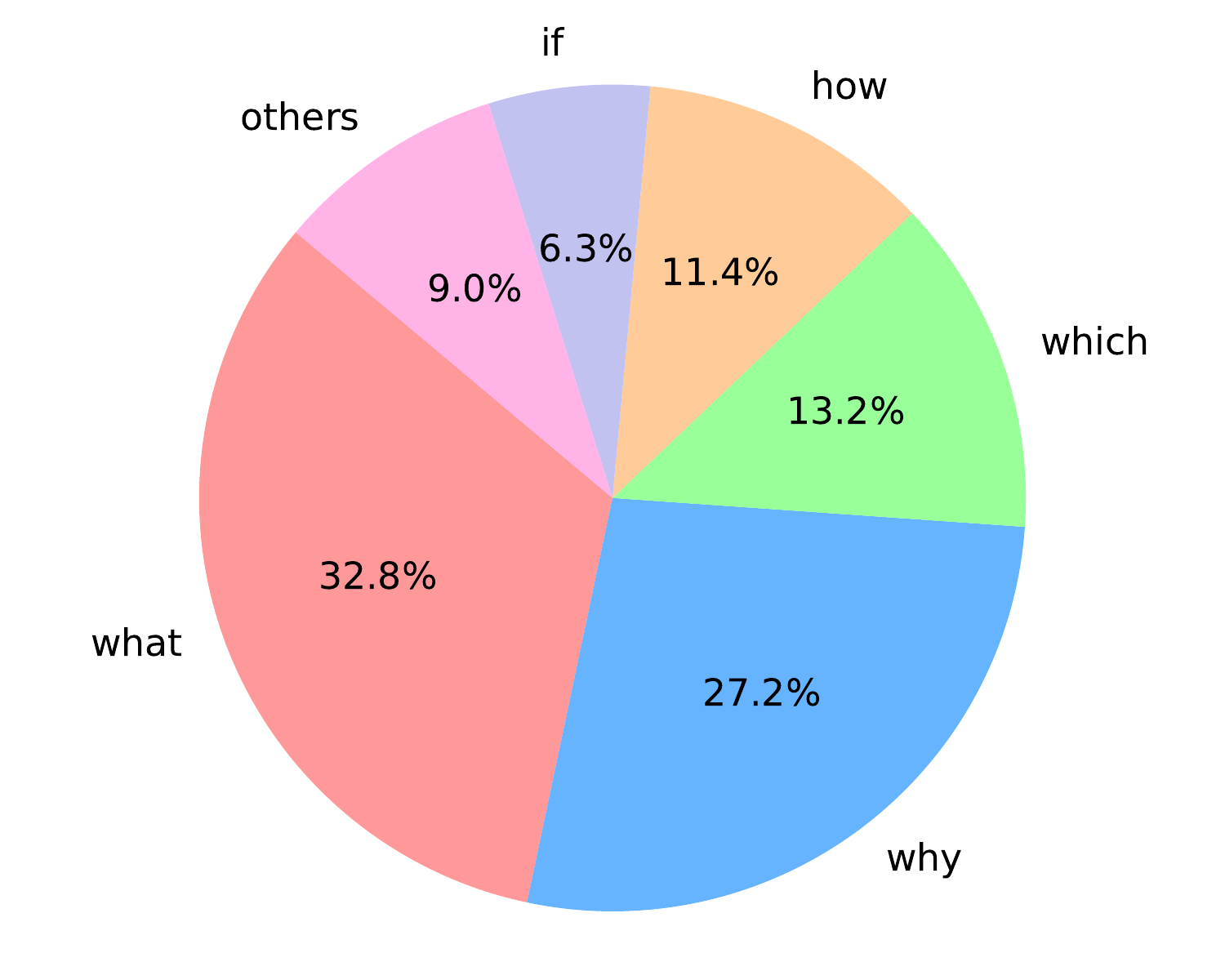}
        \caption{First word in questions}
        \label{fig:first_word_frequency}
    \end{subfigure}

    \caption{
    Statistical analysis of SIV-Bench question--answer (QA) pairs.
    (a) Average word count consistency across answer options,
    (b) distribution of correct answers among options,
    (c) question length distribution by word count,
    and (d) frequency of common first words in questions.
    }
    \label{fig:QA statistics}
\end{figure*}

To ensure the quality and balance of our SIV-Bench Question-Answer (QA) pairs, we conducted a statistical analysis of their structural properties, as summarized in Figure \ref{fig:QA statistics}.  This analysis examines several aspects: the average word count per multiple-choice option (Figure \ref{fig:average_word_count_per_option}) is relatively consistent across options A through E, minimizing length-based cues.  The distribution of correct answers (Figure \ref{fig:correct_answer_distribution}) is fairly uniform across the five options, preventing positional bias.  Furthermore, the overall question length (Figure \ref{fig:question_word_count_distribution}) peaks at around 11 words but shows a broad range, indicating variability in question complexity.  Finally, an analysis of the first word in questions (Figure \ref{fig:first_word_frequency}) reveals a diverse set of interrogative types, led by 'what,' 'why,' and 'which,' reflecting a variety of reasoning challenges posed to the models.

\subsection{Quality and Diversity Analysis} \label{appendix: qa diversity}
To further validate the quality of SIV-Bench, we conduct two analyses to ensure our benchmark encourages genuine comprehension over fitting to superficial patterns.

\subsubsection{Analysis of Template Pattern Exploitation}
SIV-Bench organizes tasks into SSU, SSR, and SDP to evaluate distinct dimensions of social cognition. To directly test whether models exploit surface-level cues, we analyze the properties of questions that a representative model (Gemini-2.0-Flash) answers correctly versus incorrectly. As shown in Table~\ref{tab:pattern_analysis_length}, while statistically significant differences in length and word count exist, the small effect sizes (Cohen’s d) indicate no meaningful structural separation between the two groups. Furthermore, Table~\ref{tab:pattern_analysis_semantic} shows a high cosine similarity between the embeddings of the correct and incorrect QA sets, along with similar internal distributions (intra-similarity and variance). This close alignment suggests that model performance is unlikely to rely on superficial statistical or semantic patterns.

\begin{table}[h!]
\centering
\caption{Length and word count comparison between correctly and incorrectly answered questions.}
\label{tab:pattern_analysis_length}
\resizebox{0.95\linewidth}{!}{%
\begin{tabular}{lcc}
\toprule
\textbf{Metric} & \textbf{Mean Length (chars)} & \textbf{Word Count} \\
\midrule
Correct  & 251.28 ± 93.88 & 42.13 ± 15.50 \\
Wrong  & 228.96 ± 94.49 & 38.21 ± 15.78 \\
t-stat              & 9.05            & 9.54 \\
p-value             & 2.36e-19        & 2.63e-21 \\
Cohen's d           & 0.2373          & 0.2516 \\
\bottomrule
\end{tabular}}
\end{table}

\begin{table}[h!]
\centering
\caption{Semantic similarity analysis between correctly and incorrectly answered question sets.}
\label{tab:pattern_analysis_semantic}
\resizebox{0.95\linewidth}{!}{%
\begin{tabular}{lc}
\toprule
\textbf{Metric} & \textbf{Value} \\
\midrule
Cosine Similarity (TF-IDF) & 0.9765 \\
Cosine Similarity (SentenceTransformer) & 0.9759 \\
\midrule
Correct Intra-similarity & 0.1689 \\
Wrong Intra-similarity & 0.1711 \\
\midrule
Correct Embedding Variance & 0.002164 \\
Wrong Embedding Variance & 0.002157 \\
\bottomrule
\end{tabular}}
\end{table}

\subsubsection{Linguistic Diversity Analysis}
The majority of questions in SIV-Bench are generated by large language models or written by human annotators, rather than using rigid templates. To objectively measure our benchmark's linguistic diversity, we compared it against several prominent video QA benchmarks using two standard metrics: \textbf{Mean Semantic Distance} (average pairwise cosine distance of sentence embeddings) and \textbf{Vector Variance} (average variance across embedding dimensions). As shown in Table~\ref{tab:diversity_comparison}, SIV-Bench exhibits high semantic diversity, ranking among the top benchmarks. These results support that SIV-Bench's questions are varied and not limited to shallow templates, thereby promoting genuine semantic understanding.

\begin{table}[h!]
\centering
\caption{Semantic diversity comparison across video QA benchmarks.}
\label{tab:diversity_comparison}
\begin{tabular}{lcc}
\toprule
\textbf{Benchmark} & \textbf{Mean Dist.} $\uparrow$ & \textbf{Variance} $\uparrow$ \\
\midrule
\textbf{SIV-Bench} & \underline{0.8321} & \underline{0.0022} \\
Video-Bench & \textbf{0.8811} & \textbf{0.0023} \\
Perception\_Test & 0.7677 & 0.0020 \\
VideoVista & 0.7604 & 0.0020 \\
Social-IQ 2.0 & 0.7260 & 0.0019 \\
MVBench & 0.7124 & 0.0019 \\
EgoSchema & 0.5411 & 0.0014 \\
\bottomrule
\end{tabular}
\end{table}

\section{Experimental Details}
\subsection{Settings Details} \label{appdx: setups}
\begin{figure*}
    \centering
    \includegraphics[width=0.8\textwidth]{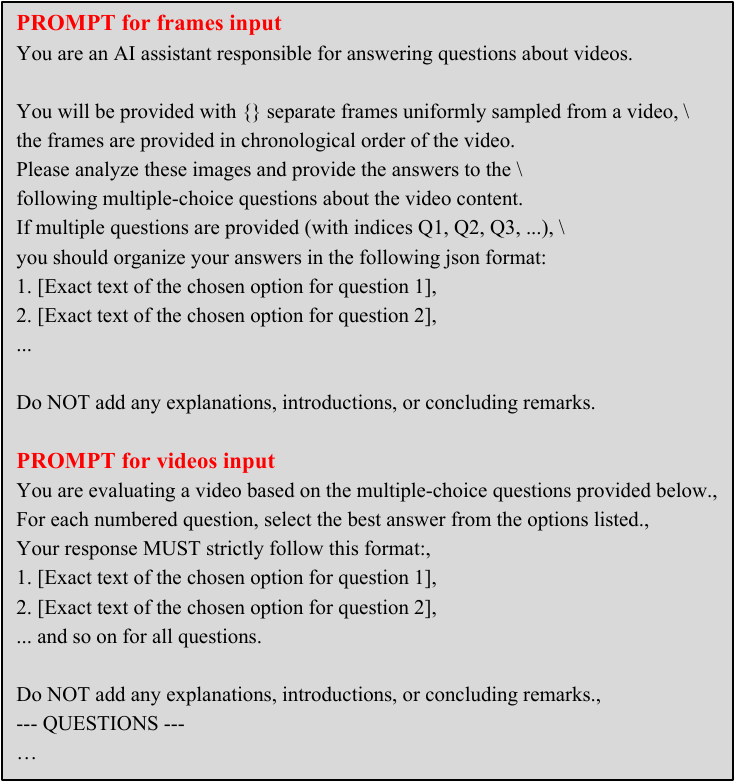}
    \caption{Standardized prompt templates used for evaluating MLLMs on SIV-Bench. Separate prompts are shown for models that process (Top) uniformly sampled frames and (Bottom) direct video input.}
    \label{fig:evaluate prompt}
\end{figure*}

This section provides further details on our experimental setup for evaluating MLLMs on SIV-Bench. Figure \ref{fig:evaluate prompt} displays the standardized prompt templates employed for model evaluations, with distinct versions tailored to models based on their input capabilities. For models that process sequences of images, the "PROMPT for frames input" (Figure \ref{fig:evaluate prompt}, Top) informs the MLLM that it will receive a set of uniformly sampled frames from a video in chronological order. For models capable of direct video processing, the "PROMPT for videos input" (Figure \ref{fig:evaluate prompt}, Bottom) is used. Both prompts clearly instruct the MLLM on its role, the task of answering multiple-choice questions based on the provided visual input, the expected JSON-like format for organizing answers (providing the exact text of the chosen option for each question), and a strict directive to avoid any extraneous text such as explanations or conversational remarks. This standardized, yet input-adaptive, prompting approach ensures consistency in task presentation across different model architectures.

For all evaluations, the specific inference parameters used for each model—such as temperature, top-p, or maximum new tokens—are adopted from their default configurations as provided within the VLMEvalKit \citep{duan2024vlmevalkit} framework. This adherence to default settings aims to reflect the out-of-the-box capabilities of these models and ensure fair comparability. The experiments are conducted on two primary compute clusters.
Cluster 1, utilized for evaluating the largest open-source models (Qwen2.5-VL-72B-Instruct and InternVL-78B), is equipped with an AMD EPYC 7642 48-Core Processor and 4x NVIDIA A100 GPUs. The total runtime for the reported experiments on this cluster is approximately 3 days.
Cluster 2, used for the remaining models, consists of an Intel(R) Xeon(R) Platinum 8369B CPU @ 2.90GHz and 8x NVIDIA RTX 3090 GPUs. The cumulative runtime for experiments on this cluster is approximately 2 days.
It should be noted that the overall research project, including preliminary testing on earlier dataset versions and exploratory experiments not included in the final results, involved a greater amount of compute time than the specific durations reported for the final benchmark evaluations.

\subsection{Failure Cases} \label{appdx: fail case}
This section presents several illustrative failure cases. We focus on examples from Gemini-2.0-Flash, a strong closed-source model, to highlight that even advanced MLLMs can fail on nuanced aspects of social perception, prediction, and reasoning. These examples are categorized by our primary assessment dimensions: SSU, SSR, and SDP, and are intended to offer concrete instances for future research and model development.

\begin{figure*}
    \centering
    \includegraphics[width=0.99\textwidth]{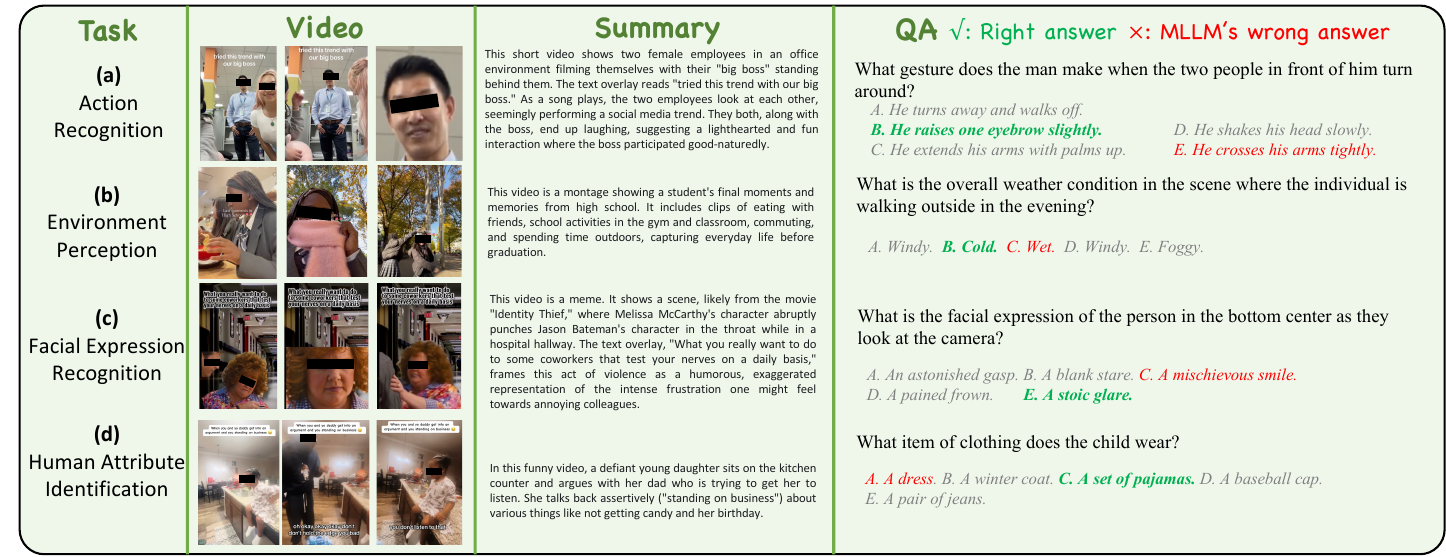}
    \caption{Examples of failure cases in Social Scene Understanding (SSU) tasks, including errors in Action Recognition, Environment Perception, Facial Expression Recognition, and Human Attribute Identification.}
    \label{fig:fail ssu}
\end{figure*}

Figure \ref{fig:fail ssu} illustrates instances where Gemini-2.0-Flash fails on SSU tasks, which require accurate perception of explicit visual elements.
\textbf{(a) In Action Recognition}, the model incorrectly identifies the man's gesture as "He crosses his arms tightly" instead of the correct "He raises one eyebrow slightly", missing a subtle but distinct facial action.
\textbf{(b) For Environment Perception}, when asked about the weather, the model failed to capture the details of the characters in the scene wearing thick scarves and down jackets to infer that the correct answer was "cold", but instead wrongly chose "wet".
\textbf{(c) In Facial Expression Recognition}, the model describes the expression as "A mischievous smile" rather than the correct "A stoic glare", misinterpreting the nuanced facial expression display.
\textbf{(d) For Human Attribute Identification}, concerning the child's clothing, the model selects "A dress" instead of the correct "A set of pajamas", failing to correctly identify common apparel.

\begin{figure*}
    \centering
    \includegraphics[width=0.99\textwidth]{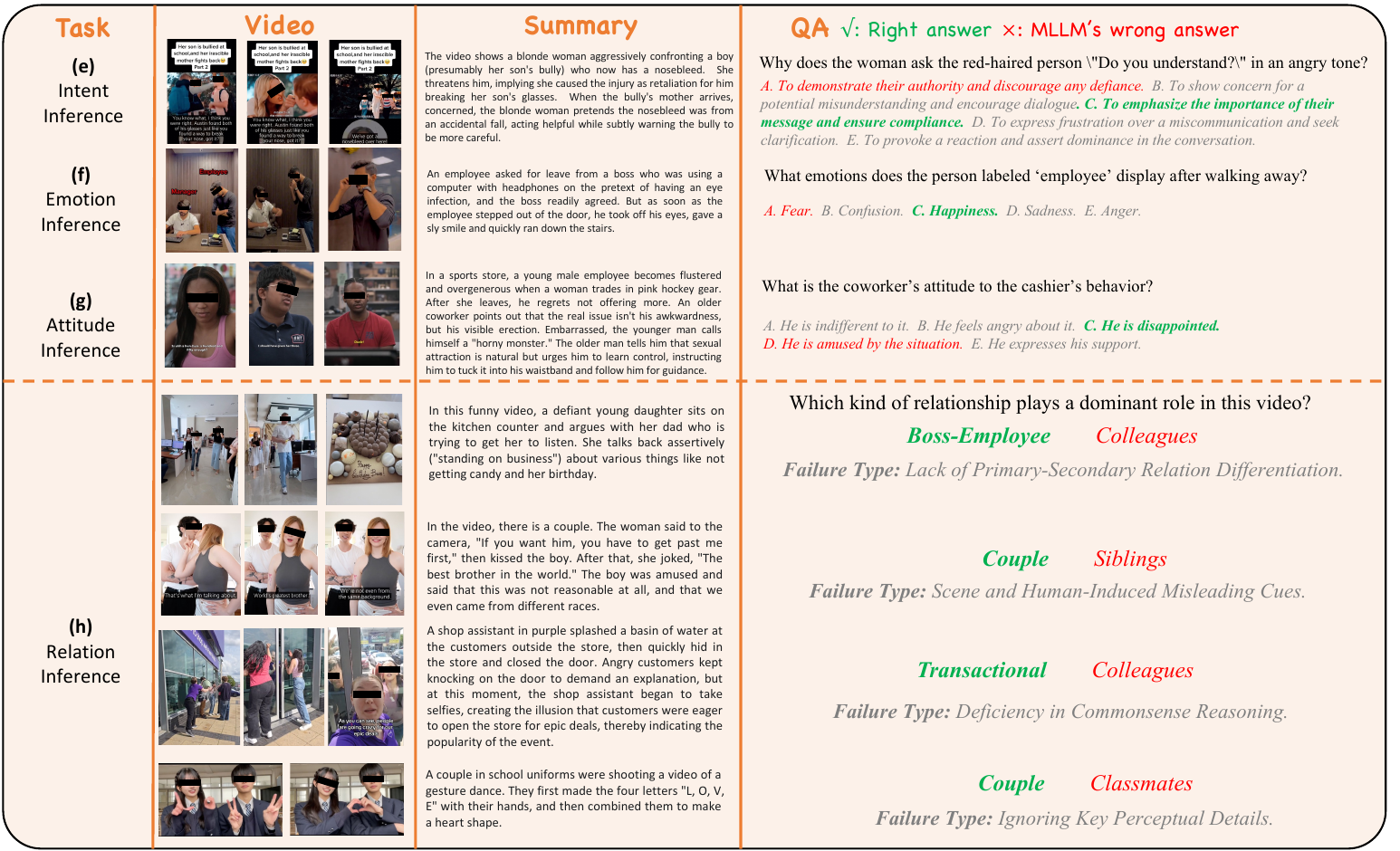}
    \caption{Examples of failure cases in Social State Reasoning (SSR) tasks, highlighting difficulties in Intent Inference, Emotion Inference, Attitude Inference, and Relation Inference.}
    \label{fig:fail ssr}
\end{figure*}

Figure \ref{fig:fail ssr} presents failure cases of Gemini-2.0-Flash on SSR tasks, which involve inferring unobservable mental states and relationships.
\textbf{(e) In Intent Inference}, when a woman says "do you understand?" to a boy who bullies her son in an angry tone, it is to teach him a lesson and warn him not to bully her son again, not for "discourage any defiance", because in fact no child much younger than her can form defiance against her.
\textbf{(f) For Emotion Inference}, this employee is happy instead of scared after leaving because he successfully deceives the boss into giving him a vacation.
\textbf{(g) In Attitude Inference}, the coworker is dissatisfied and disappointed with the cashier's nervousness, panic and even physical reactions when seeing female customers. This could also be seen from his subsequent warning to the cashier not to do so anymore.
\textbf{(h) For Relation Inference}, we present case studies on the failure patterns of the four common models listed in the main text in this task.

\begin{figure*}
    \centering
    \includegraphics[width=0.99\textwidth]{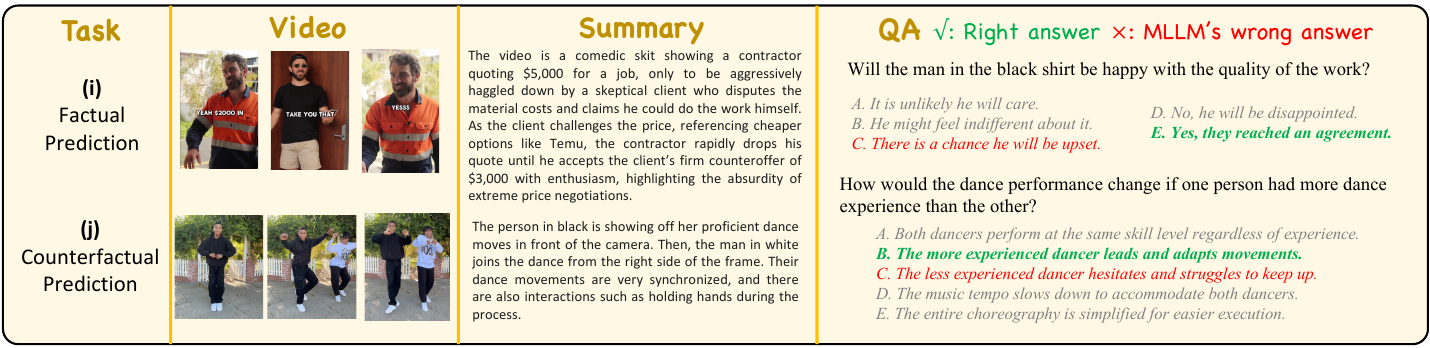}
    \caption{Examples of failure cases in Social Dynamics Prediction (SDP) tasks, covering both Factual Prediction and Counterfactual Prediction.}
    \label{fig:fail sdp}
\end{figure*}

Figure \ref{fig:fail sdp} highlights errors made by Gemini-2.0-Flash in SDP tasks, which require predicting future events or reasoning about hypothetical scenarios.
\textbf{(i) In Factual Prediction}, when asked if the person in the black shirt would be satisfied with the workers' work, since the two of them have already reached an agreement with smiles at the end of the video, it could be inferred that the answer was "yes", but the model chooses another answer.
\textbf{(j) For Counterfactual Prediction}, the video shows the dance interaction between a mother and her son. The question raised is what would happen if one of them had more dance experience. This can be inferred from the positive and relaxed interaction between the two in the video. The most likely answer is "The more experienced dancer leads and adapts movements". For example, a son leads his mother to learn dancing happily, rather than "The less experienced dancer hesitates and struggles to keep up". They have a good relationship, and the probability of negative performance like "hesitates" and "struggles" is lower.

\subsection{Text-Only Control for Shortcut Analysis}
\label{appdx:text_only_control}

To further quantify the extent to which \ours{} can be solved from textual priors alone, we introduce a \textbf{text-only control} that removes all visual and audio information and provides models with only the question and the five answer options. This setting is designed as a direct probe of \emph{textual shortcut solvability}: if a substantial portion of the benchmark could be solved without access to the video, then models should retain strong performance under this control.

\paragraph{Setup.}
We evaluate five representative MLLMs under the text-only setting, including both closed- and open-source systems. In this control, models receive the same question and candidate options as in the standard evaluation, but \emph{no video frames, no audio, and no subtitle input} are provided. Since each question in \ours{} has five candidate answers, the random-guessing baseline is $20\%$. We report accuracy on the full benchmark and on the three task categories, namely Social Scene Understanding (SSU), Social State Reasoning (SSR), and Social Dynamics Prediction (SDP).

\paragraph{Results.}
As shown in Table~\ref{tab:text_only_control_all}, text-only performance remains close to the random baseline across all five models, and the ordering of models is notably less stable than in the standard multimodal setting. This suggests that, once visual and audio evidence are removed, models largely lose the consistent advantage observed in the full benchmark. We also observe that SSU does not consistently become easier in the text-only setting, which is expected since many SSU questions require direct grounding in visual content. Meanwhile, some SSR and SDP items remain marginally solvable from textual priors alone, but the gains are still small and far below the standard multimodal setting. Moreover, the proportion of questions answered correctly by all five models, or by at least four out of five models, stays very low. These findings indicate that \ours{} contains relatively few questions that can be reliably solved from textual priors alone, and they further support our claim that strong performance on the more reasoning-intensive portions of \ours{}, especially SSR and SDP, cannot be adequately explained by text-only shortcuts.

\begin{table*}[t]
\centering
\small
\setlength{\tabcolsep}{5pt}
\begin{tabular}{lccccc}
\toprule
\textbf{Model / Indicator} & \textbf{Overall} & \textbf{SSU} & \textbf{SSR} & \textbf{SDP} & $\Delta$ \textbf{vs Rand.} \\
\midrule
Gemini-2.5-Pro (text-only)   & 24.0 & 23.0 & 22.0 & 26.0 & +4.0 \\
GPT-4o (text-only)       & 23.0 & 21.0 & 24.0 & 22.0 & +3.0 \\
Qwen2.5-VL-7B (text-only)   & 22.0 & 24.0 & 20.0 & 21.0 & +2.0 \\
InternVL3-8B (text-only) & 25.0 & 20.0 & 23.0 & 29.0 & +5.0 \\
LLaVA-Video (text-only)   & 21.0 & 19.0 & 22.0 & 23.0 & +1.0 \\
\midrule
Random (5-way)           & 20.0 & 20.0 & 20.0 & 20.0 & +0.0 \\
\midrule
All 5/5 models correct (\%)      & 0.7 & 0.9 & 0.6 & 0.8 & -- \\
At least 4/5 models correct (\%) & 4.2 & 5.8 & 3.6 & 4.9 & -- \\
\bottomrule
\end{tabular}
\caption{Text-only control and dataset-level shortcut indicators on \ours{}. In the text-only setting, models receive only the question and the five answer options, without any visual, audio, or subtitle input. The results remain close to random guessing, and the low cross-model agreement further suggests that only a small fraction of questions can be solved from textual priors alone.}
\label{tab:text_only_control_all}
\end{table*}

\subsection{Analysis of Chain-of-Thought Prompting}

To investigate the impact of explicit reasoning on model performance, we conducted a preliminary experiment using a Chain-of-Thought (CoT) prompting strategy. We prepended the instruction \textit{"Let's think step by step. First, output your reasoning process, and then output the final answer."} to our standard evaluation prompt. The overall accuracy on the \texttt{origin} videos, with and without CoT, is presented in Table~\ref{tab:cot_results}.

The results indicate that applying a generic CoT prompt did not yield significant performance improvements for most models. For several smaller open-source models (e.g., mPLUG-Owl3, LLaVA-Video), it resulted in a notable performance decrease. We observed that this is often because these models struggle to consistently adhere to the more complex two-stage output format (i.e., providing reasoning before the final answer in the required format), leading to failures in our answer parsing logic.

The primary goal of SIV-Bench is to establish a fair, consistent, and reproducible evaluation of baseline model capabilities. Our standardized prompting strategy, which aligns with widely used toolkits like VLMEvalKit, ensures this fairness. While techniques like CoT are powerful for eliciting maximum performance from certain capable models (e.g., Gemini-2.5-Pro), introducing them as a default can create a confounding variable. Such a setup might shift the evaluation from testing inherent social reasoning to testing complex instruction-following abilities. Therefore, our main experiments use a direct-answering prompt to maintain a level playing field. Nonetheless, these findings suggest that developing more specialized reasoning methods tailored to social intelligence is a valuable direction for future work that builds upon this benchmark.

\begin{table}[h!]
\centering
\caption{Comparison of Overall Accuracy (\%) on the \texttt{origin} videos with and without Chain-of-Thought (CoT) prompting.}
\label{tab:cot_results}
\resizebox{0.48\textwidth}{!}{
\begin{tabular}{lcc}
\toprule
\textbf{Model} & \textbf{Acc (Origin)} & \textbf{Acc (CoT)} \\
\midrule
mPLUG-Owl3              & 42.06 & 41.79 \\
LLaVA-OneVision         & 41.97 & 41.22 \\
LLaVA-Video             & 41.09 & 40.13 \\
Qwen2.5-VL-7B-Instruct  & 44.02 & 44.76 \\
InternVL-8B             & 45.82 & 44.99 \\
\midrule
Qwen2.5-VL-72B-Instruct & 58.80 & 59.13 \\
InternVL-78B            & 55.46 & 55.25 \\
\midrule
o4-mini                 & 55.68 & 55.79 \\
GPT-4o                  & 58.02 & 57.91 \\
Gemini-2.0-Flash        & 56.40 & 56.53 \\
Gemini-2.5-Flash        & 57.87 & 57.71 \\
Gemini-2.5-Pro          & 61.65 & 61.77 \\
\bottomrule
\end{tabular}
}
\end{table}

\subsection{Statistical Significance Analysis}
\label{app:statistical_analysis}

To validate the reliability of our comparative claims, we conducted McNemar's tests on the full dataset ($N=5,455$). This paired non-parametric test is appropriate for comparing the performance of two classifiers on the same dataset.

\paragraph{Model Ranking.} 
We verified the leadership of our SOTA model. The performance difference between the top-performing \textbf{Gemini-2.5-Pro} (61.65\%) and the second-best model \textbf{Qwen2.5-VL-72B} (58.80\%) is highly statistically significant ($p < 0.001$), confirming the robustness of the leaderboard rankings.

\paragraph{Task Difficulty.} 
We confirmed that the performance stratifications across our three core dimensions are not due to chance. For Gemini-2.5-Pro, the performance gaps between \textbf{SSU} (85.07\%) and \textbf{SDP} (60.45\%), as well as between \textbf{SDP} and \textbf{SSR} (54.30\%), are all highly statistically significant ($p < 0.001$).

\paragraph{Subtitle Influence.} 
We performed significance testing on the subtitle conditions (Table 4). Our analysis reveals that the minor \textit{Overall} improvement from adding subtitles (`+sub`) is not statistically significant ($p = 0.12$). However, the impact is task-dependent: the negative impact of removing text (`-sub`) is statistically significant for the \textbf{SSR} task ($p < 0.05$), and the benefit of added subtitles (`+sub`) is significant for the \textbf{SDP} task ($p < 0.05$).

\section{SIV-Bench-Hard Details}
\label{app:siv_bench_hard}

To rigorously evaluate the upper limits of current MLLMs, establish a robust human baseline, and probe the reasoning processes beyond simple accuracy, we curated and analyzed a challenging subset of our dataset, termed \textbf{SIV-Bench-Hard}. This section details the setup, human performance, and a multi-dimensional analysis of model reasoning quality on this subset.

\subsection{Experimental Setup}
\textbf{Dataset Curation.} We selected a subset of 200 questions from the human-generated portion of SIV-Bench. These questions were specifically chosen for their complexity and reliance on deep social understanding, filtering out items that could be solved via superficial visual cues.

\begin{figure}
    \centering
    \includegraphics[width=0.98\linewidth]{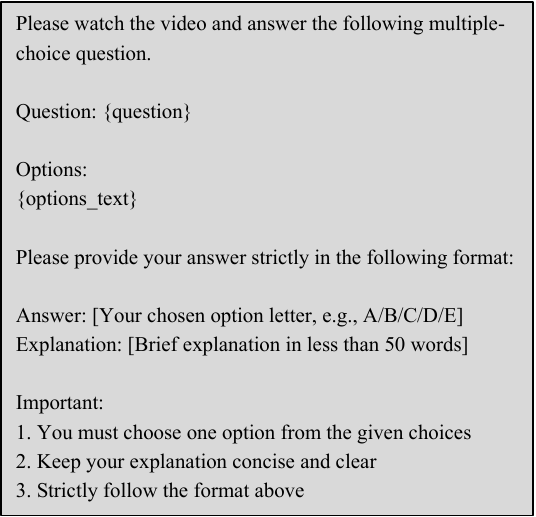}
    \caption{Model prompt for SIV-Bench-Hard.}
    \label{fig:test_hard}
\end{figure}

\begin{figure}
    \centering
    \includegraphics[width=0.98\linewidth]{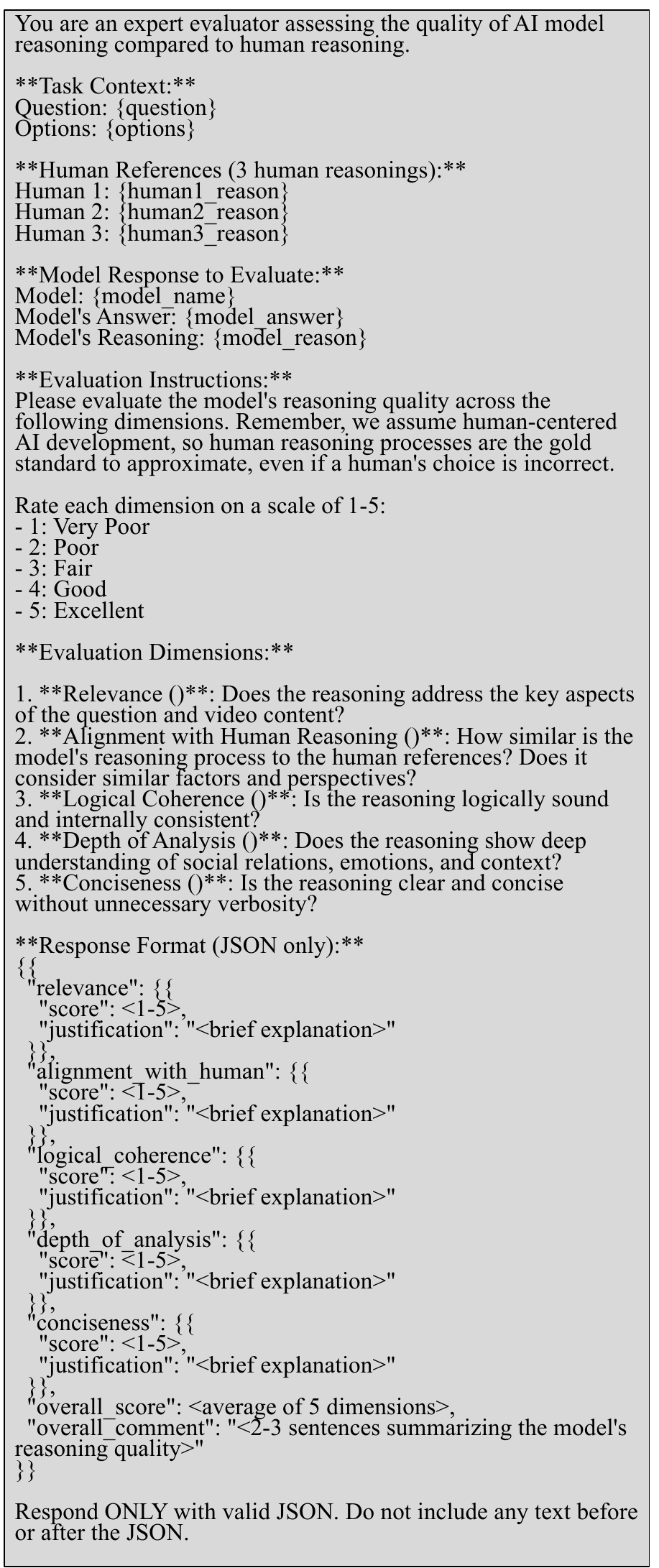}
    \caption{LLM-judge prompt for scoring explanations in SIV-Bench-Hard.}
    \label{fig:evaluate_hard}
\end{figure}

\textbf{Task Definition.} Unlike the standard multiple-choice evaluation, this study required both human annotators and MLLMs to provide: (1) the selected answer option, and (2) a free-text \textit{reasoning explanation} justifying their choice. The prompt is shown in Figure \ref{fig:test_hard} and \ref{fig:evaluate_hard}. This allows for a deeper examination of the cognitive process.

\textbf{Participants.} We recruited 3 independent human annotators to perform this task to establish a human baseline. We evaluated a suite of state-of-the-art MLLMs, including Gemini-3-Pro, GPT-5.1, Gemini-2.5-Pro, Gemini-2.5-Flash, Qwen2.5-VL-7B, and GPT-4o-mini.

\subsection{Quantitative Analysis of Reasoning}
To quantify the divergence observed in the qualitative scores, we performed an embedding-based analysis. We encoded all reasoning texts (both human and model) using the \texttt{paraphrase-multilingual-MiniLM-L12-v2} model.

\textbf{Distinguishability.} We trained a Random Forest classifier to distinguish between human and model explanations based on their embeddings. The classifier achieved an accuracy of \textbf{92.17\%} (compared to a 50\% random baseline). This high classification accuracy indicates that the latent semantic features of model reasoning are fundamentally distinct from those of humans.

\textbf{Statistical Significance.} We further validated this difference using Mann-Whitney U tests across the embedding dimensions. The tests confirmed that the distributions of human and model embeddings are significantly different ($p < 0.05$) on 8 out of 10 principal dimensions. Collectively, these results provide quantitative evidence that current MLLMs, despite their linguistic fluency, employ reasoning processes that are statistically distinguishable from human social cognition.

The SIV-Bench-Hard subset, along with the human reasoning annotations and analysis code, will be released to facilitate future research into bridging this gap.

\subsection{Correlation Between Answer Correctness and Reasoning Fidelity}
\label{sec:reasoning_fidelity}

To rigorously verify that model performance is grounded in genuine social reasoning rather than superficial visual shortcuts (e.g., background cues or object co-occurrence), we analyze the relationship between the correctness of a model's answer and the semantic quality of its reasoning trace. We hypothesize that if models were merely "guessing" via shortcuts, their generated explanations would lack alignment with human cognitive processes even when they fortuitously select the correct option.

We perform an embedding-based analysis on the \textit{SIV-Bench-Hard} subset using the \texttt{paraphrase-multilingual-MiniLM-L12-v2} model. For each of the six evaluated models, we calculate the cosine similarity between the model's generated reasoning and the human expert ground truth. The results are stratified based on whether the model answers the multiple-choice question correctly or incorrectly.

\begin{figure*}[t]
    \centering
    \includegraphics[width=0.98\textwidth]{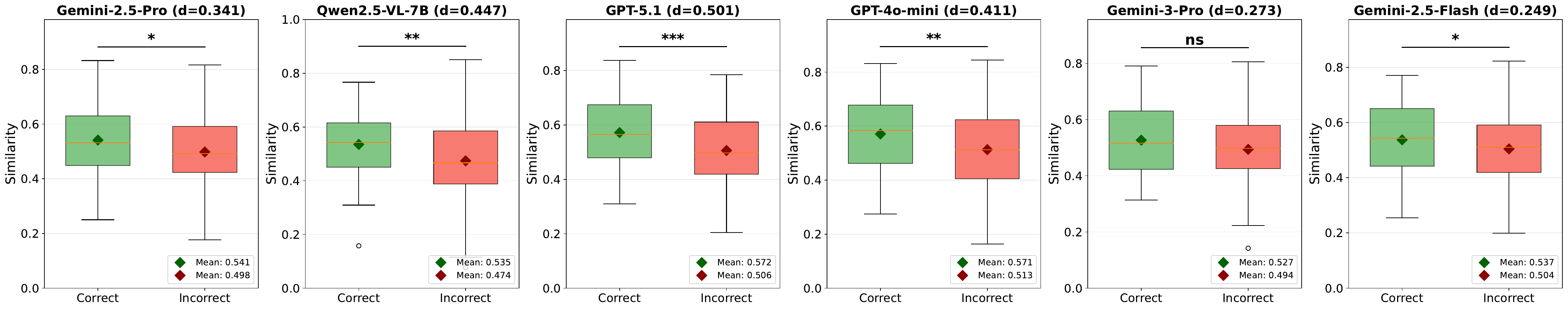}
    \caption{Box plots illustrating the distribution of semantic similarity scores between model reasoning and human ground truth, stratified by answer correctness. We observe a consistent trend where reasoning traces for correct answers exhibit significantly higher alignment with human social cognition compared to incorrect answers. Statistical significance is denoted by * ($p<0.05$), ** ($p<0.01$), and *** ($p<0.001$).}
    \label{fig:reasoning_similarity}
\end{figure*}

\begin{table}[h]
    \centering
    \caption{Quantitative comparison of reasoning similarity to human ground truth between correct and incorrect responses. The \textbf{Gap ($\Delta$)} represents the increase in similarity when the model answers correctly. Significance is calculated using the Mann-Whitney U Test.}
    \label{tab:reasoning_similarity_stats}
    \resizebox{0.48\textwidth}{!}{
    \begin{tabular}{lcccc}
        \toprule
        \textbf{Model} & \textbf{\makecell{Similarity \\ (Correct)}} & \textbf{\makecell{Similarity \\ (Incorrect)}} & \textbf{Gap ($\Delta$)} & \textbf{\makecell{Significance \\ ($p$-value)}} \\
        \midrule
        GPT-5.1 & 0.572 & 0.506 & +0.066 & \textbf{$<$ 0.001} (***) \\
        Qwen2.5-VL-7B & 0.535 & 0.474 & +0.061 & \textbf{$<$ 0.01} (**) \\
        GPT-4o-mini & 0.571 & 0.513 & +0.058 & \textbf{$<$ 0.01} (**) \\
        Gemini-2.5-Pro & 0.541 & 0.498 & +0.043 & \textbf{$<$ 0.05} (*) \\
        Gemini-2.5-Flash & 0.537 & 0.504 & +0.033 & \textbf{$<$ 0.05} (*) \\
        Gemini-3-Pro & 0.527 & 0.494 & +0.033 & 0.067 (n.s.) \\
        \bottomrule
    \end{tabular}
    }
\end{table}

As illustrated in Figure~\ref{fig:reasoning_similarity} and summarized in Table~\ref{tab:reasoning_similarity_stats}, we observe a statistically significant positive gap in similarity scores across 5 out of the 6 models. Notably, GPT-5.1 demonstrates the strongest effect, with a similarity gap of 0.066 ($p < 0.001$). 

These findings provide empirical evidence that correctness in SIV-Bench is strongly correlated with human-like social reasoning. The significant degradation in reasoning alignment during failure cases suggests that models do not rely on "guessing" via superficial cues; rather, successful performance necessitates a cognitive process that mirrors human social understanding.

\section{Broader Impact} \label{appdx: impact}
SIV-Bench is designed to foster positive advancements in artificial social intelligence, potentially leading to more empathetic, context-aware, and collaborative AI systems for beneficial applications such as assistive technologies, improved human-AI teaming, and richer content understanding. However, enhancing AI's grasp of social dynamics also presents risks. These capabilities could be misused for sophisticated manipulation, disinformation, or invasive surveillance, and unaddressed biases in data could be amplified, leading to inequitable outcomes. We offer SIV-Bench as a research tool to transparently assess MLLM capabilities and limitations in the social domain, thereby encouraging the community to proactively consider these ethical challenges and develop robust safeguards alongside continued innovation in social AI.

\section{Use of Ai Assistants}
The LLM was utilized exclusively to aid and polish the writing, including for tasks such as improving
grammar and clarity, refining sentence structure, and ensuring stylistic consistency

\end{document}